\documentclass[runningheads]{llncs}
\usepackage{graphicx}
\usepackage{comment}
\usepackage{amsmath,amssymb} %
\usepackage{color}
\usepackage{epsfig}
\usepackage{shortbold}
\usepackage{booktabs}
\usepackage{hyperref}
\usepackage{xcolor}

\usepackage[width=122mm,left=12mm,paperwidth=146mm,height=193mm,top=12mm,paperheight=217mm]{geometry}

\begin{document}
\pagestyle{headings}
\mainmatter
\def\ECCVSubNumber{2313}  %

\title{Associative3D: Volumetric Reconstruction from \\Sparse Views} %

\titlerunning{Associative3D: Volumetric Reconstruction from Sparse Views}

\author{Shengyi Qian$^*$ \and
Linyi Jin$^*$ \and
David F. Fouhey}
\authorrunning{S. Qian et al.}
\institute{
University of Michigan\\
\email{\{syqian,jinlinyi,fouhey\}@umich.edu}
}
\maketitle

{\let\thefootnote\relax\footnotetext{{$^*$ equal contribution}}}

\begin{abstract}
  This paper studies the problem of 3D volumetric reconstruction
  from two views of a scene with an unknown camera. While seemingly
  easy for humans, this problem poses many challenges for computers
  since it requires simultaneously reconstructing objects in the two
  views while also figuring out their relationship. We propose a new approach
  that estimates reconstructions, distributions over the
  camera/object and camera/camera transformations, as well as an
  inter-view object affinity matrix. This information is then jointly reasoned
  over to produce the most likely explanation of the scene. We train and
  test our approach on a dataset of indoor scenes, and rigorously
  evaluate the merits of our joint reasoning approach. 
  Our experiments show that it is able to recover reasonable scenes from sparse views, while the problem is still challenging.
  Project site: \url{https://jasonqsy.github.io/Associative3D}.
\end{abstract}

\newcommand{\papername}{Associative3D}

\definecolor{msgreenline}{rgb}{0.304,0.367,0.175}
\definecolor{msredline}{rgb}{0.376,0.157,0.151}
\definecolor{msblueline}{rgb}{0.155,0.253,0.371}
\definecolor{msorangeline}{rgb}{0.484,0.294,0.137}
\definecolor{mspurpleline}{rgb}{0.251,0.196,0.318}

\definecolor{msgreen}{rgb}{0.608,0.733,0.349}
\definecolor{msred}{rgb}{0.753,0.314,0.302}
\definecolor{msblue}{rgb}{0.310,0.506,0.741}
\definecolor{msorange}{rgb}{0.969,0.588,0.275}
\definecolor{mspurple}{rgb}{0.502,0.392,0.635}

\begin{figure}[h]
  \vspace{-3em}
  \centering
  \includegraphics[width=0.95\textwidth]{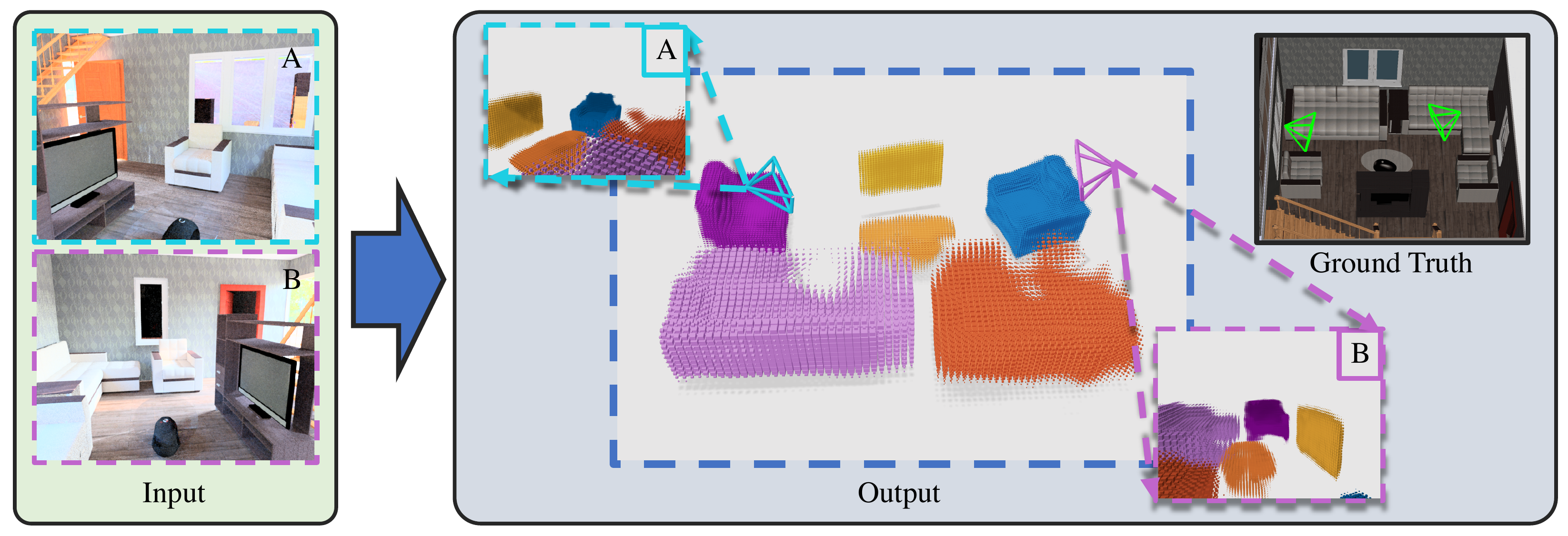}
  \caption{Given two views from unknown cameras, we aim 
  to extract a coherent 3D space in terms of 
  a set of volumetric objects placed in the scene. 
  We represent the scene with a factored representation \cite{tulsiani2018factoring} that splits the scene
  into per-object voxel grids with a scale and pose.  }
  \label{fig:teaser}
  \vspace{-2em}
\end{figure}

\section{Introduction}
\label{sec:introduction}
How would you make sense of the scene in Fig. \ref{fig:teaser}? After rapidly
understanding the individual pictures, one can fairly quickly attempt to match
the objects in each: the TV on the left in image A must go with the TV on
the right in image B, and similarly with the couch. Therefore, the two
chairs, while similar, are not actually the same object. Having pieced this
together, we can then reason that the two images depict the same scene, but
seen with a 180$^\circ$ change of view and infer the 3D structure of
the scene. Humans have an amazing ability to reason about the 3D structure
of scenes, even with as little as two sparse views with an unknown relationship.
We routinely use this ability to understand images taken at an event,
look for a new apartment on the Internet, or evaluate possible hotels (e.g.,
for ECCV). The goal of this paper is to give the same ability to computers.

Unfortunately, current techniques are not up to this challenge of volumetric reconstruction given two views from unknown cameras: this
approach requires both reconstruction and pose estimation.
Classic methods based on correspondence \cite{Hartley04,Crandall2013pami}
require many more views in practice and cannot make inferences about unseen
parts of the scene (i.e., what the chair looks like from behind) since this
requires some form of learning. While there has been success in learning-based
techniques for this sort of object reconstructions \cite{Choy16,Girdhar16b,tulsiani2018factoring,kulkarni20193d}, it is unknown how to
reliably stitch together the set of reconstructions into a single coherent
story. Certainly there are systems that can identify pose with respect to 
a fixed scene \cite{kendall2015posenet} or a pair of views
\cite{en2018rpnet}; these approaches, however cannot reconstruct.

This paper presents a learning-based approach to this problem, whose results are shown in
Fig.~\ref{fig:teaser}. The system can take two views with unknown relationship, and
produce a 3D scene reconstruction for both images jointly. This 3D scene reconstruction
comprises a set of per-object reconstructions rigidly placed in the scene with a pose as in
\cite{tulsiani2018factoring,kulkarni20193d,Li2019}. Since the 3D scene reconstruction
is the union of the posed objects, getting the 
{\it 3D scene reconstruction} correct requires getting both the {\it 3D object reconstruction} right
as well as correctly identifying {\it 3D object pose}.
Our key insight is that jointly reasoning about objects and poses improves the
results. Our method, described in Section \ref{sec:approach}, predicts evidence
including: (a) voxel reconstructions for
each object; (b) distributions over rigid body
transformations between cameras and objects; and (c) an
inter-object affinity for stitching. Given this
evidence, our system can stitch them together to find the
most likely reconstruction. 
As we empirically demonstrate in Section \ref{sec:experiments}, this joint reasoning
is crucial -- understanding each image independently and then 
estimating a relative pose performs substantially worse compared to our approach. 
These are conducted
on a challenging and large dataset of indoor scenes. %
We also show some common failure modes and demonstrate transfer to NYUv2~\cite{Silberman12} dataset.

Our primary contributions are: (1) Introducing a novel problem -- volumetric scene reconstruction from two unknown sparse views; (2) Learning an inter-view object affinity to find correspondence between images; (3) Our joint system, including the stitching stage, %
is better than adding individual components.

\section{Related Work}
\label{sec:related}
The goal of this work is to take two views from cameras related by an unknown
transformation and produce a single volumetric understanding of the scene. 
This touches on a number of important problems in computer vision ranging from
the estimation of the pose of objects and cameras, full shape of objects, and
correspondence across views. Our approach deliberately builds heavily on these
works and, as we show empirically, our success depends crucially on their
fusion.

This problem poses severe challenges for classic correspondence-based approaches
\cite{Hartley04}. From a purely geometric perspective, we are totally out of luck:
even if we can identify the position of the camera via epipolar geometry and
wide baseline stereo \cite{Pritchett98,Mishkin15}, we
have no correspondence for most objects in Fig. \ref{fig:teaser} that would permit
depth given known baseline, let alone another view that would help lead to
the understanding of the full shape of the chair.%

Recent work has tackled identifying this full volumetric reconstruction via
learning. Learning-based 3D has made significant progress recently, including 2.5D representations~\cite{eigen2015predicting,wang2015designing,chen2019learning,lasinger2019towards}, single object reconstruction~\cite{wu2017marrnet,zhang2018learning,groueix2018,richter2018matryoshka,choy20163d}, and scene understanding~\cite{chen2019holisticpp,huang2018holistic,liu2018floornet,liu2019planercnn,du2018learning}.
Especially, researchers have developed increasingly detailed volumetric
reconstructions beginning with objects
\cite{Choy16,Girdhar16b,Gkioxari2019} and then moving to scenes
\cite{tulsiani2018factoring,kulkarni20193d,Li2019,nie2020total3dunderstanding} as a composition of object
reconstructions that have a pose with respect to the camera. Focusing on full volumetric reconstruction, our approach builds on this progression, and creates an
understanding that is built upon jointly reasoning over parses of two scenes, affinities,
and relative poses; as we empirically show, this produces improvements in results.
Of these works, we are most inspired by Kulkarni et al.~\cite{kulkarni20193d} in that
it also reasons over a series of relative poses; our work builds on top of this
as a base inference unit and handles multiple images. 
We note that while we
build on a particular approach to scenes \cite{kulkarni20193d} and objects
\cite{Girdhar16b}, our approach is general. 

While much of this reconstruction work is single-image, some is
multiview, although usually in the case of an isolated object
\cite{kar2017learning,Choy16,huang2018deep} or with hundreds of views
\cite{huang2018deepmvs}. Our work aims at the particular task of as little as
two views, and reasons over multiple objects. 
While traditional local features \cite{lowe2004distinctive} are insufficient to support reasoning over objects, 
semantic features are useful \cite{duggal2019deeppruner,wang2018multi,bowman2017probabilistic}.

At the same time, there has been considerable progress in identifying the
relative pose from images \cite{melekhov2017relative,kendall2015posenet,en2018rpnet,bao2012semantic}, RGB-D Scans \cite{Yang2019,yang2020extreme} or video sequences \cite{zhou2017unsupervised,salas2013slam++,sui2020geofusion}.
Of these, our work is most related to
learning-based approaches to identifying relative pose from RGB images, and semantic Structure-from-Motion \cite{bao2012semantic} and SLAM \cite{salas2013slam++}, which make use of semantic elements to improve the estimation of camera pose.
We build upon this work in our approach, especially work
like RPNet \cite{en2018rpnet} that directly predicts relative pose, although we do so 
with a regression-by-classification formulation that provides uncertainty. 
As we show empirically, propagating this uncertainty forward lets us reason
about objects and produce superior results to only focusing on pose.

\section{Approach}
\label{sec:approach}
\begin{figure*}[!t]
    \centering
    \includegraphics[width=\textwidth]{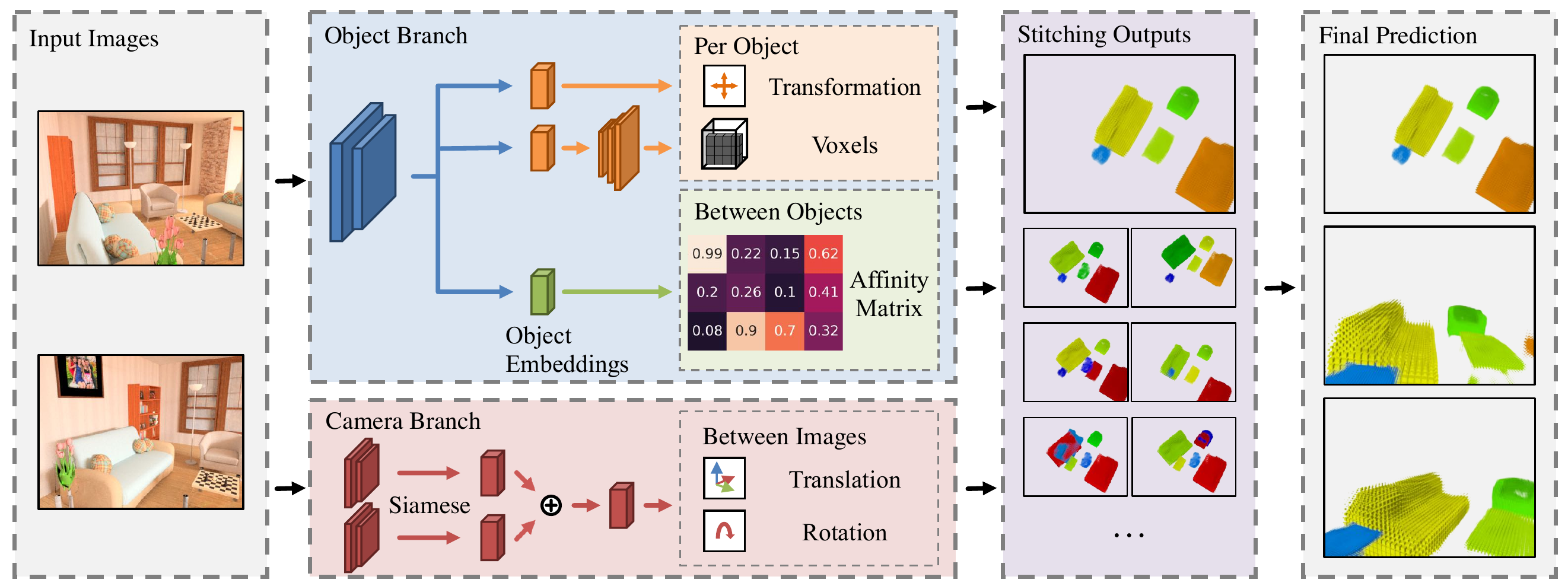}
    \caption{{\bf Our approach.} We pass the two RGB image inputs into
    two branches that extract evidence, which is then fused together
    to stitch a final result. Our first network, \textcolor{msblue}{\bf object branch}, is a detection
    network following \cite{kulkarni20193d} that produces a set of 
    \textcolor{msorange}{\bf objects} in terms of voxels and a transformation
    into the scene. We also predict an object embedding
    which we can use to form an \textcolor{msgreen}{\bf affinity matrix} between objects across images.
    Our second network, \textcolor{msred}{\bf camera branch}, is a siamese network that 
    predicts a distribution over translations and rotations between the
    cameras. Finally our \textcolor{mspurple}{\bf stitching stage} examines the evidence from the networks
    and produces a final prediction.
} \label{fig:pipeline}
\end{figure*}

The goal of the system is to map a pair of sparse views of a room to a full 3D
reconstruction. 
As input, we assume a pair of images of a room. As output, we produce a set of objects represented as voxels,
which are rigidly transformed and anisotropically scaled into the scene
in a single coordinate frame. We achieve this with an approach, summarized in Fig. \ref{fig:pipeline},
that consists of three main parts: an object branch, a camera branch, and a stitching stage.

The output space is a factored representation of a 3D scene, similar to
\cite{tulsiani2018factoring,kulkarni20193d,Li2019}. Specifically, in contrast to using
a single voxel-grid or mesh, the scene is represented as a set of per-object voxel-grids 
with a scale and pose that are placed in the scene. These can be converted to a single
3D reconstruction by taking their union, and so improving the 3D reconstruction can be done
by either improving the per-object voxel grid or improving its placement in the scene.

The first two parts of our approach are two neural networks. An \textbf{object branch} examines each
image and detects and produces single-view 3D reconstructions for objects in the
camera's coordinate frame, as well as a 
per-object embedding that helps find the object in the other image.
At the same time, an \textbf{camera branch}
predicts relative pose between images, represented as a {\it distribution} over a discrete 
set of rigid transformations between the cameras. These networks are trained separately
to minimize complexity.

The final step, a \textbf{stitching stage}, combines these together. 
The output
of the two networks gives: a collection of objects per image in the image's coordinate frame;
a cross-image affinity 
which predicts object correspondence in two views;
and a set of likely transformations from one camera to the other.
The stitching stage aims to select a final set of predictions
minimizing an objective function that aims to ensure that similar objects are in the same location,
the camera pose is likely, etc. Unlike the first two stages, this is an optimization
rather than a feedforward network.

\subsection{Object Branch}

The goal of our object branch is to take an image and produce a set of reconstructed
objects in the camera's coordinate frame as well as an embedding that lets us
match across views. We achieve this by extending 3D-RelNet \cite{kulkarni20193d} and adjust
it as little as possible to ensure fair comparisons. We refer the reader for a
fuller explanation in \cite{kulkarni20193d,tulsiani2018factoring}, but briefly, these
networks act akin to an object detector like Faster-RCNN \cite{ren2015faster} with additional
outputs. As input, 3D-RelNet takes as input an image and a set of 2D bounding box
proposals, and maps the image through convolutional layers to a feature map, from
which it extracts per-bounding box convolutional features. These features
pass through fully connected layers to predict: a detection score (to suppress bad proposals), voxels (to represent the
object), and a transformation to the world frame (represented by rotation,
scale, and translation and calculated via both per-object and pairwise poses).
We extend this to also produce an n-dimensional embedding $\eB \in \mathbb{R}^n$ 
on the unit sphere (i.e., $||\eB||_2^2 = 1$) that helps associate objects across images. 

We use and train the embedding by creating a cross-image affinity matrix
between objects.  
Suppose the first and second images have $N$ and $M$ objects
each with embeddings $\eB_i$ and $\eB_j'$ respectively. We then define our 
affinity matrix $\AB \in \mathbb{R}^{N \times M}$ as 
\begin{equation}
    \AB_{i,j} = \sigma(k \eB_i^T \eB_j')
    \label{eqn:aff}
\end{equation}
where $\sigma$ is the sigmoid/logistic function and where $k= 5$ scales the output.
Ideally, $A_{i,j}$ should indicate whether objects $i$ and $j$ are the
same object seen from a different view, where $A_{i,j}$ is high
if this is true and low otherwise.

We train this embedding network using ground-truth bounding box proposals
so that we can easily calculate a ground-truth affinity matrix
$\hat{\AB}$. We then minimize $L_{\mathrm{aff}}$, a balanced mean-square loss between $\AB$ 
and $\hat{\AB}$: if all positive labels are $(i, j) \in \mathcal{P}$, and all
negative labels are $(i, j) \in \mathcal{N}$, then the loss is
\begin{equation}
    L_{\mathrm{aff}} = \frac{1}{|\mathcal{P}|}\sum_{(i,j)\in \mathcal{P}} (A_{ij} - \hat{A}_{ij})^2 + \frac{1}{|\mathcal{N}|}\sum_{(i,j)\in \mathcal{N}} (A_{ij} - \hat{A}_{ij})^2.
\end{equation}
which balances positive and negative labels (since affinity
is imbalanced).

\subsection{Camera Branch}

Our camera branch aims to identify or narrow down the possible relationship between
the two images. We approach this by building a siamese network \cite{bromley1994signature} that predicts the relative camera pose $T_c$ between the two images.  
We use ResNet-50 \cite{he2016deep} to extract features from two input images. 
We concatenate the output features and then use two linear layers to predict the translation and rotation.

We formulate prediction of rotation and translation as a classification
problem to help manage the uncertainty in the problem. We found that propagating uncertainty
(via top predictions) was helpful: a single feedforward network suggests
likely rotations and a subsequent stage can make a more detailed assessment
in light of the object branch's predictions. Additionally, even if we care
about only one output, we found regression-by-classification to be helpful
since the output tended to have multiple modes (e.g., being
fairly certain of the rotation modulo $90^\circ$ by recognizing that both images depict 
a cuboidal room). Regression tends to split the difference, producing predictions which satisfy
neither mode, while classification picks one, as observed in \cite{tulsiani2018factoring,Ladicky14b}.

We cluster the rotation and translation vectors into 30 and 60 bins respectively, and predict two multinomial distributions over them.
Then we minimize the cross entropy loss.
At test time, we select the cartesian product of the top 3 most likely bins for
rotation and top 10 most likely bins for translation as the final prediction
results. The results are treated as proposals in the next section.

\subsection{Stitching Object and Camera Branches}

Once we have run the object and camera branches, our goal is to then produce a single stitched result. As input to this step, our
object branch gives: for view 1, with $N$ objects, the voxels $V_1, \ldots, V_N$ and transformations $T_1, \ldots, T_N$;
and similarly, for $M$ objects in view 2, the voxels $V_1', \ldots, V_M'$ and transformations $T_1', \ldots, T_M'$; and
a cross-view affinity matrix $\AB \in [0,1]^{N \times M}$. Additionally,
we have a set of potential camera transformations $P_1,\ldots,P_F$ between two views. 

The goal of this section is to integrate this evidence to find a final
cross-camera pose $P$ and correspondence $\CB \in \{0,1\}^{M
\times N}$ from view $1$ to view $2$. This correspondence is one-to-one and has the
option to ignore an object (i.e., $\CB_{i,j} = 1$ if and only if $i$ and $j$
are in correspondence and for all $i$, $\sum_j \CB_{i,j} \le 1$, and similarly
for $\CB^T$).

We cast this as a minimization problem over $P$ and $\CB$ including terms in the objective function that incorporate the above evidence.
The cornerstone term is one that integrates all the evidence to examine the quality of the stitch, akin to trying and seeing
how well things match up under a camera hypothesis. We implement this by computing the distance $\mathcal{L}_D$
between corresponding object voxels according to $\CB$, once the transformations are applied, or:
\begin{equation}
    \mathcal{L}_D =  \frac{1}{|\CB|_1} \sum_{(i,j)~\textrm{s.t.}~\CB_{i,j}=1} D(P(T_i(V_i)), T'_j(V_j')).
\end{equation}
Here, $D$ is the chamfer distance between points on the edges of each shape, as defined in \cite{price2019inferring,sharma2018csgnet}, or for two point clouds $X$ and $Y$:
\begin{equation}
    D(X, Y) = \frac{1}{|X|} \sum_{x\in X} \min_{y\in Y} ||x-y||_2^2 + \frac{1}{|Y|} \sum_{y\in Y} \min_{x\in X} ||x-y||_2^2.
\end{equation}
Additionally, we have terms that reward making $\CB$ likely according to our object and image networks, or:
the sum of similarities between corresponding objects according to the affinity matrix $\AB$, 
$\mathcal{L}_S = \sum_{(i,j), \CB_{i,j} = 1} (1-A_{i,j})$; as well as 
the probability of the camera pose transformation $P$ from the image network $\mathcal{L}_P = (1-Pr(P))$. Finally,
to preclude trivial solutions, we include a term rewarding minimizing the number of un-matched objects, or $\mathcal{L}_U = \min(M,N)-|\CB|_1$.
In total, our objective function is the sum of these terms, or:
\begin{equation}
    \min_{P,{\CB}}\quad \mathcal{L}_D + \lambda_P \mathcal{L}_P + \lambda_S \mathcal{L}_S +  \lambda_U \mathcal{L}_U.
\end{equation}

The search space is intractably large, so we optimize the objective function by RANSAC-like search over the top 
hypotheses for $P$ and feasible object correspondences. 
For each top hypothesis of $P$, we randomly sample $K$ object correspondence proposals. Here we use $K=128$. It is generally sufficient since the correspondence between two objects is feasible only if the similarity of them is higher than a threshold according to the affinity matrix. We use random search over object correspondences because the search space increases factorially between the number of objects in correspondence. Once complete, we average the translation and scale, and randomly pick one rotation and shape from corresponding objects. 
Averaging performs poorly for rotation since there are typically multiple rotation modes that cannot be averaged: a symmetric table is correct at either 0$^\circ$ or 180$^\circ$ but not at 90$^\circ$.
Averaging voxel grids does not make sense since there are partially observable objects.
We therefore pick one mode at random for rotation and shape.
Details are available in the appendix.

\section{Experiments}
\label{sec:experiments}

We now describe a set of experiments that aim to address the following questions:
(1) how well does the proposed method work and are there simpler approaches
that would solve the problem? and (2) how does the method solve the problem?
We first address question (1) by evaluating the proposed approach compared 
to alternate approaches both qualitatively and by evaluating the full
reconstruction quantitatively.
We then address question (2) by evaluating individual components of the system. 
We focus on what the affinity matrix learns and whether the stitching stage can jointly improve object correspondence and relative camera pose estimation.
Throughout, we test our approach on the SUNCG dataset \cite{song2017semantic,zhang2017physically}, following 
previous work \cite{tulsiani2018factoring,kulkarni20193d,Li2019,Yang2019,zhang2017physically}. 
To demonstrate
transfer to other data, we also show qualitative results on NYUv2 \cite{Silberman12}.

\begin{figure*}[!t]
    \centering
    
    \begin{tabular}{cccccccc}
        \toprule
    
    \multicolumn{2}{c}{Input Images} & \multicolumn{2}{c}{Camera 1} &\multicolumn{2}{c}{Camera 2} & \multicolumn{2}{c}{Birdview} \\
    Image 1 & Image 2 & Prediction & GT & Prediction & GT & Prediction & GT\\
    \midrule
    \frame{\includegraphics[width=0.11\textwidth]{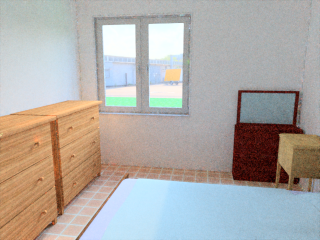}}
    & \frame{\includegraphics[width=0.11\textwidth]{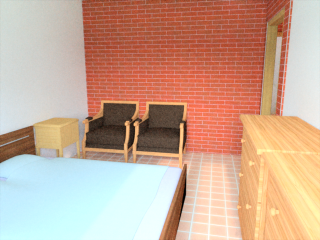}}
    & \frame{\includegraphics[width=0.11\textwidth]{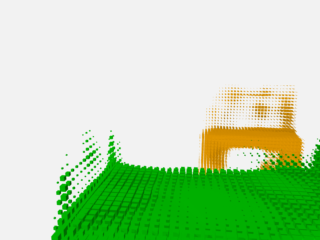}}
    & \frame{\includegraphics[width=0.11\textwidth]{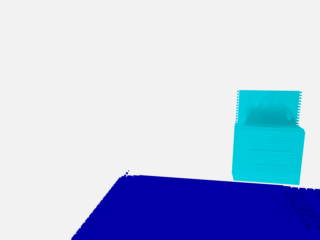}}
    & \frame{\includegraphics[width=0.11\textwidth]{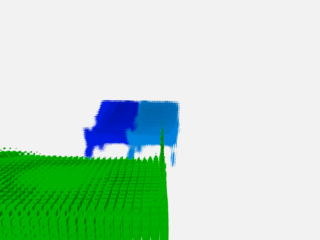}}
    & \frame{\includegraphics[width=0.11\textwidth]{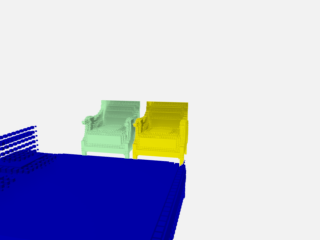}}
    & \frame{\includegraphics[width=0.11\textwidth]{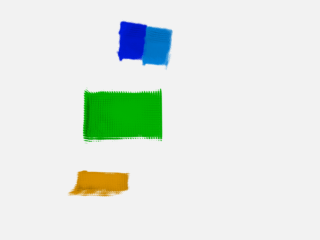}}
    & \frame{\includegraphics[width=0.11\textwidth]{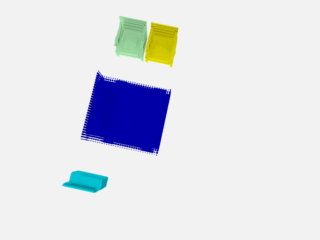}}\\

    \frame{\includegraphics[width=0.11\textwidth]{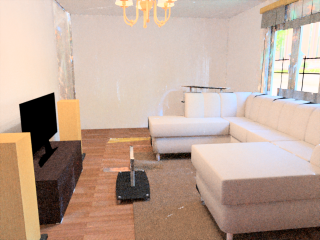}}
    & \frame{\includegraphics[width=0.11\textwidth]{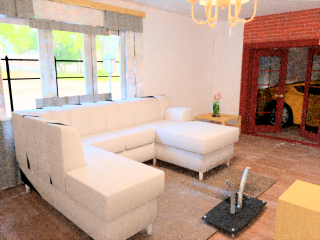}}
    & \frame{\includegraphics[width=0.11\textwidth]{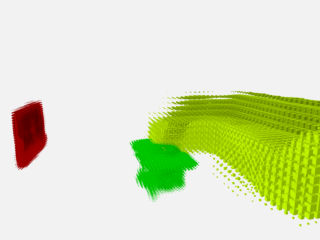}}
    & \frame{\includegraphics[width=0.11\textwidth]{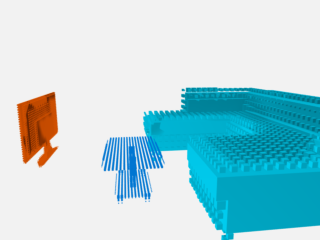}}
    & \frame{\includegraphics[width=0.11\textwidth]{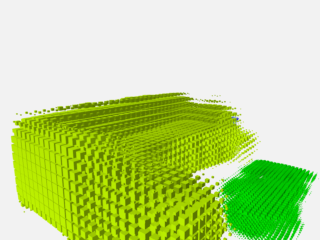}}
    & \frame{\includegraphics[width=0.11\textwidth]{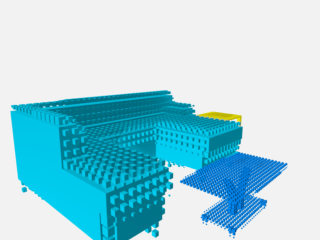}}
    & \frame{\includegraphics[width=0.11\textwidth]{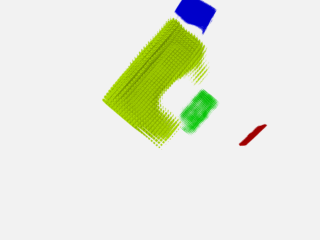}}
    & \frame{\includegraphics[width=0.11\textwidth]{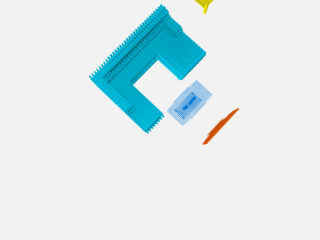}}\\
    
    \frame{\includegraphics[width=0.11\textwidth]{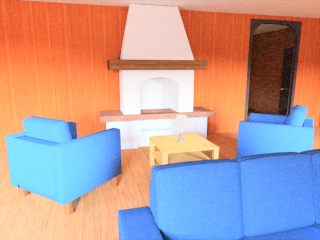}}
    & \frame{\includegraphics[width=0.11\textwidth]{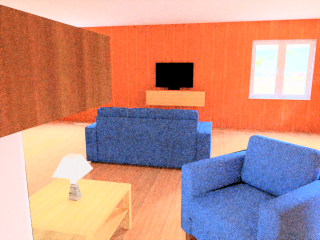}}
    & \frame{\includegraphics[width=0.11\textwidth]{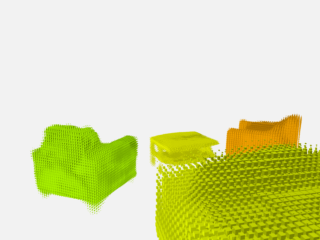}}
    & \frame{\includegraphics[width=0.11\textwidth]{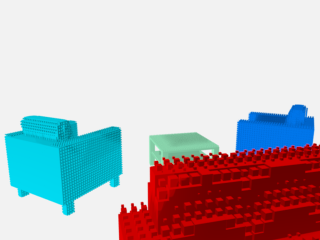}}
    & \frame{\includegraphics[width=0.11\textwidth]{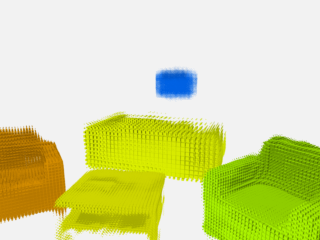}}
    & \frame{\includegraphics[width=0.11\textwidth]{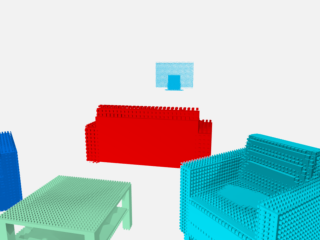}}
    & \frame{\includegraphics[width=0.11\textwidth]{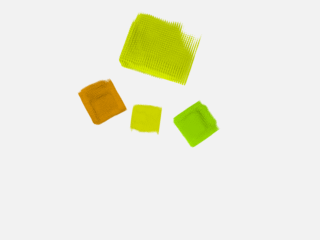}}
    & \frame{\includegraphics[width=0.11\textwidth]{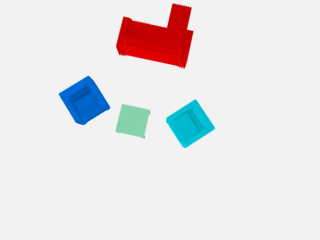}}\\
    
    \frame{\includegraphics[width=0.11\textwidth]{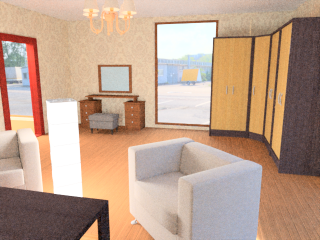}}
    & \frame{\includegraphics[width=0.11\textwidth]{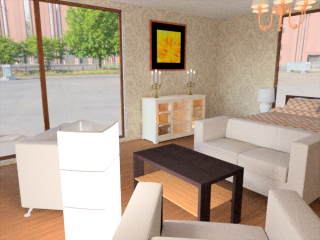}}
    & \frame{\includegraphics[width=0.11\textwidth]{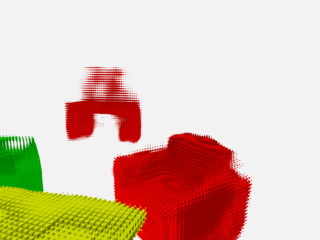}}
    & \frame{\includegraphics[width=0.11\textwidth]{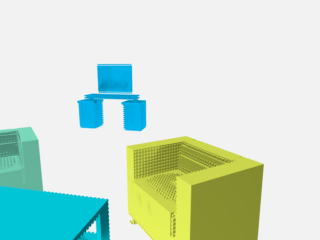}}
    & \frame{\includegraphics[width=0.11\textwidth]{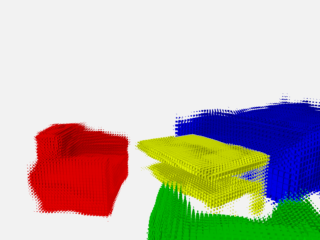}}
    & \frame{\includegraphics[width=0.11\textwidth]{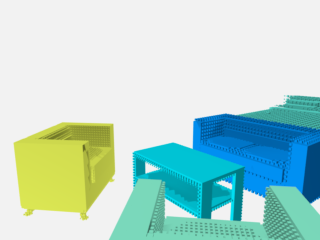}}
    & \frame{\includegraphics[width=0.11\textwidth]{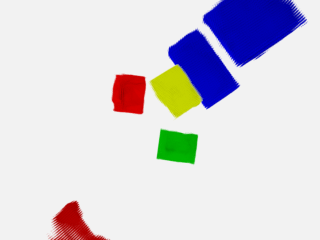}}
    & \frame{\includegraphics[width=0.11\textwidth]{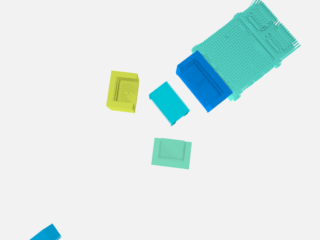}}\\
    
    \frame{\includegraphics[width=0.11\textwidth]{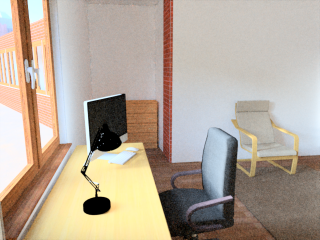}}
    & \frame{\includegraphics[width=0.11\textwidth]{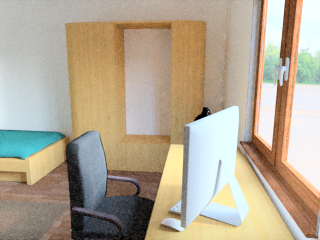}}
    & \frame{\includegraphics[width=0.11\textwidth]{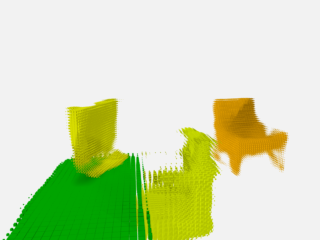}}
    & \frame{\includegraphics[width=0.11\textwidth]{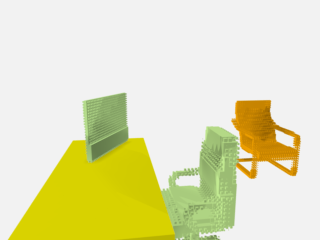}}
    & \frame{\includegraphics[width=0.11\textwidth]{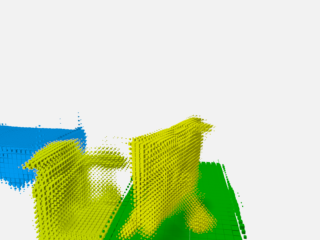}}
    & \frame{\includegraphics[width=0.11\textwidth]{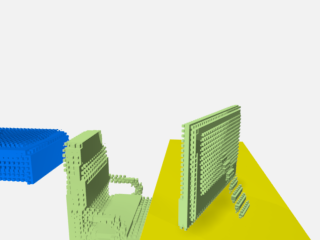}}
    & \frame{\includegraphics[width=0.11\textwidth]{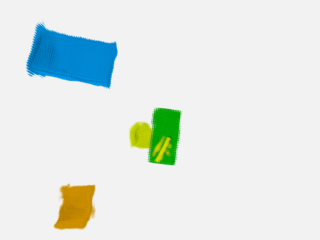}}
    & \frame{\includegraphics[width=0.11\textwidth]{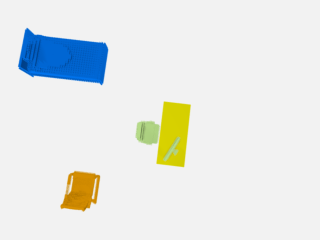}}\\

    \bottomrule
    \end{tabular}
    \caption{Qualitative results on the SUNCG test set \cite{song2017semantic}. The final 3D predictions are shown in three different camera poses (1) the same camera as image 1; (2) the same camera as image 2; (3) a bird view to see all the objects in the whole scene. In the prediction, red/orange objects are from the left image, blue objects are from the right image, green/yellow objects are stitched.}
    \label{fig:suncg-wall}
\end{figure*}

\subsection{Experimental Setup}

We train and do extensive
evaluation on SUNCG \cite{song2017semantic} since it provides 3D scene ground truth including voxel representation of objects.
There are realistic datasets such as ScanNet \cite{dai2017scannet} and Matterport3D \cite{chang2017matterport3d}, but they only provide non-watertight meshes. Producing filled object voxel representation from non-watertight meshes remains an open problem.
For example, Pix3D \cite{sun2018pix3d} aligns IKEA furniture models with images, but not all objects are labeled.

\vspace{0.75em}
\noindent
\textbf{Datasets.} 
We follow the
70\%/10\%/20\% training, validation and test split of houses from \cite{kulkarni20193d}.
For each house, we randomly sample up to ten rooms; for each room, we
randomly sample one pair of views. Furthermore, we filter the
validation and test set: we eliminate pairs where there is no overlapping
object between views, and pairs in which all of one image's objects are 
in the other view (i.e., one is a proper subset of the other). 
We do not filter the training set since learning relative pose requires a large and diverse training
set.
Overall, we have 247532/1970/2964 image pairs for training, validation
and testing, respectively. Following \cite{tulsiani2018factoring}, we use six object classes - bed, chair, desk, sofa, table and tv.

\vspace{0.75em}
\noindent
{\bf Full-Scene Evaluation:} 
Our output is a full-scene reconstruction, represented as a set of per-object voxel grids that are posed and scaled in the scene.
A scene prediction can be totally wrong if one of the objects has correct shape while its translation is off by 2 meters.
Therefore, we quantify performance by treating the problem as a 3D detection problem in which we predict a series of 3D boxes and voxel grids.
This lets us evaluate which aspect of the problem currently hold methods back.
Similar to \cite{kulkarni20193d}, for each object, we define error metrics as follows:

\noindent
\textbullet\ Translation ($\mathbf{t}$): Euclidean distance, or $\delta_t = ||t - \hat{t}||_2$, thresholded at $\delta_t = 1$m.

\noindent
\textbullet\ Scale ($\mathbf{s}$): Average log difference in scaling factors, or $\delta_s = \frac{1}{3} \sum_{i=1}^3 |\log_2(s_1^i) - \log_2(s_2^i)|$, thresholded at $\delta_s = 0.2$.

\noindent
\textbullet\ Rotation ($\mathbf{R}$): Geodesic rotation distance, or $\delta_q = (2)^{-1/2} ||\log(\mathbf{R}^T\hat{\mathbf{R}})||_F$, thresholded at $\delta_q = 30^{\circ}$.

\noindent
\textbullet\ Shape ($\mathbf{V}$): 
Following~\cite{tatarchenko2019single}, we use F-score@0.05 to measure the difference between prediction and ground truth, thresholded at $\delta_V = 0.25$.

A prediction is a true positive only
if all errors are lower than our thresholds. 
We calculate the precision-recall curve based on that and report average precision (AP). 
We also report AP for each single error metric.

\vspace{0.75em}
\noindent
\textbf{Baselines.} 
Since there is no prior work on this task, our experiments compare to ablations and alternate forms of our method.
We use the following baseline methods, each of which tests a concrete hypothesis.
({\bf Feedforward}):
This method uses the object branch to recover single-view 3D scenes,
and our camera branch to estimate the relative pose between different views. We
ignore the affinity matrix and pick the top-1 relative pose predicted by the
camera branch. There can be many duplicate objects in the output of this approach. This tests if
a simple feedforward method is sufficient.
({\bf NMS}): In addition to the feedforward approach, we perform non-maximum
suppression on the final predictions. If two objects are close to each other,
we merge them. This tests if a simple policy to merge objects would work.
({\bf Raw Affinity}): Here, we use the predicted affinity matrix to merge objects
based on top-1 similarity from the affinity matrix. This tests whether our stitching stage is necessary.
({\bf \papername}): This is our complete method. We optimize the objective function by searching possible rotations, translations and object correspondence.

\begin{figure}[!t]
    \centering
    \scriptsize
    \begin{tabular}{c@{\hskip3pt}c@{\hskip8pt}c@{\hskip3pt}c@{\hskip3pt}c@{\hskip3pt}c@{\hskip8pt}c}
    \toprule
    
    Image 1 & Image 2 & Feedforward & NMS & Raw Affinity & {\bf Ours} & GT\\
    \midrule
    \frame{\includegraphics[width=0.12\textwidth]{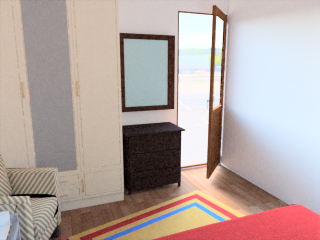}}
    & \frame{\includegraphics[width=0.12\textwidth]{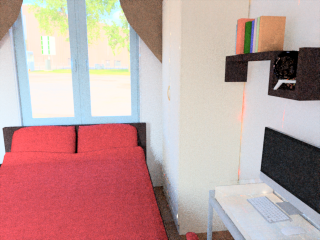}}
    & \frame{\includegraphics[width=0.12\textwidth]{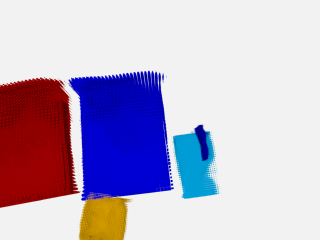}}
    & \frame{\includegraphics[width=0.12\textwidth]{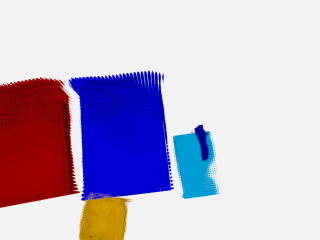}}
    & \frame{\includegraphics[width=0.12\textwidth]{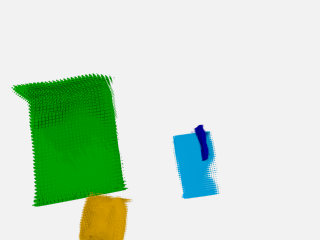}}
    & \frame{\includegraphics[width=0.12\textwidth]{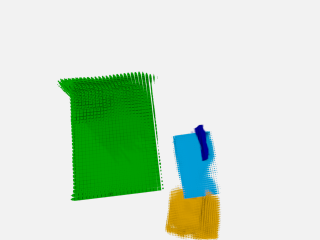}}
    & \frame{\includegraphics[width=0.12\textwidth]{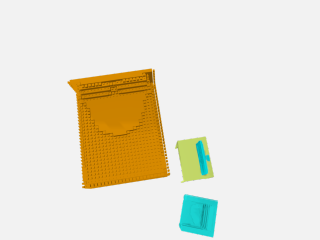}}\\

    \frame{\includegraphics[width=0.12\textwidth]{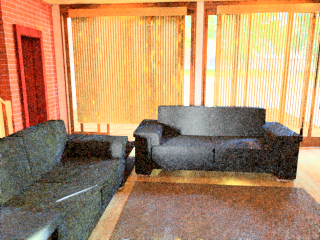}}
    & \frame{\includegraphics[width=0.12\textwidth]{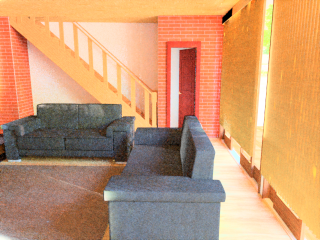}}
    & \frame{\includegraphics[width=0.12\textwidth]{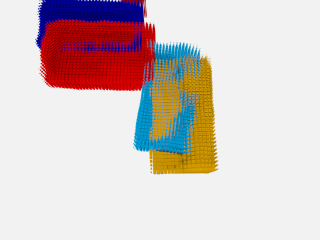}}
    & \frame{\includegraphics[width=0.12\textwidth]{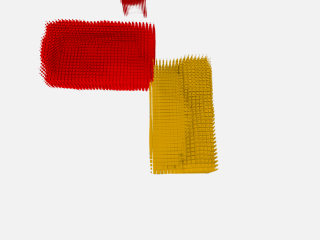}}
    & \frame{\includegraphics[width=0.12\textwidth]{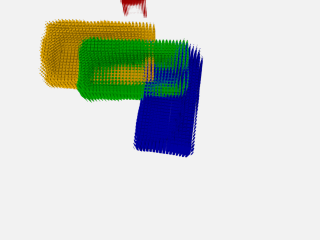}}
    & \frame{\includegraphics[width=0.12\textwidth]{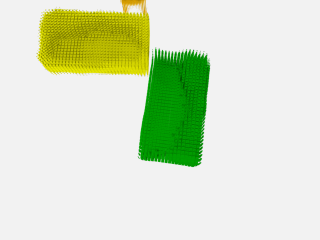}}
    & \frame{\includegraphics[width=0.12\textwidth]{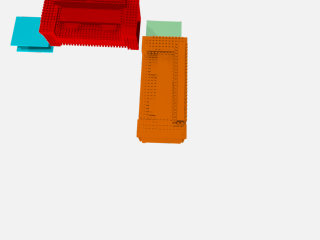}}\\

    \frame{\includegraphics[width=0.12\textwidth]{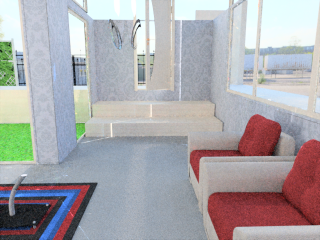}}
    & \frame{\includegraphics[width=0.12\textwidth]{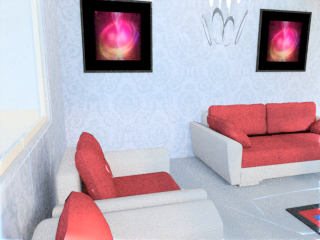}}
    & \frame{\includegraphics[width=0.12\textwidth]{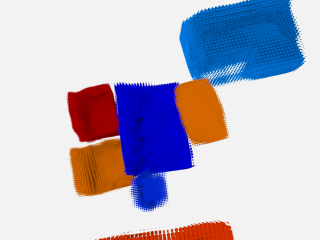}}
    & \frame{\includegraphics[width=0.12\textwidth]{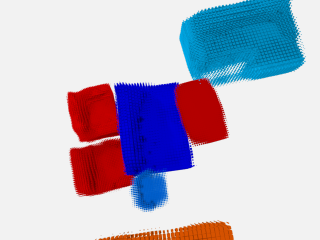}}
    & \frame{\includegraphics[width=0.12\textwidth]{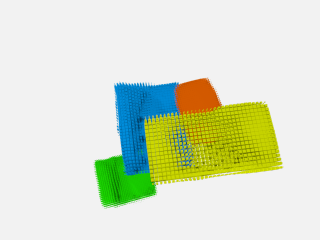}}
    & \frame{\includegraphics[width=0.12\textwidth]{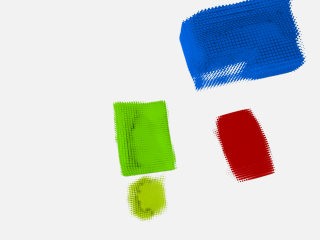}}
    & \frame{\includegraphics[width=0.12\textwidth]{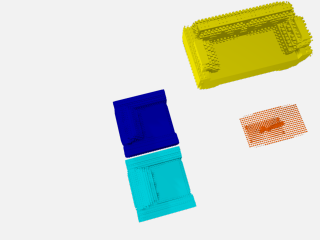}}\\

    \bottomrule
    \end{tabular}
    \caption{Comparison between {\papername} and alternative approaches. \textbf{Row 1:} {\papername} fixes the incorrect top-1 relative camera pose in light of a single bed in the room. \textbf{Row 2}: NMS works when the relative camera pose is accurate. \textbf{Row 3:} {\papername} outperforms all alternative approaches in finding correspondence in object clutter.}
    \label{fig:comparison}
\end{figure}

\subsection{Full Scene Evaluation}
\label{sec:expfull}

We begin by evaluating our full scene reconstruction. Our output is a set of
per-object voxels that are posed and scaled in the scene. The quality of reconstruction of a single object is decided by both the voxel grids and the object pose.

First, we show qualitative examples
from the proposed method in Fig. \ref{fig:suncg-wall} as well as a comparison 
with alternate approaches in Fig. \ref{fig:comparison} on the SUNCG test set. 
The Feedforward approach tends to have duplicate objects since it does not know object correspondence.
However, figuring out the camera pose and common objects is a
non-trivial task. 
Raw Affinity does not work since it may merge objects based on their similarity, regardless of possible global conflicts.
NMS works when the relative camera pose is accurate but cannot work when many objects are close to each other.
Instead, {\papername} demonstrates the ability to jointly reason over reconstructions, object pose and camera pose to produce a reasonable explanation of the scene. More qualitative examples are available in
the supplementary material.

We then evaluate our proposed approach quantitatively. In a factored representation~\cite{tulsiani2018factoring}, both {\it object} poses and shapes are equally important
to the full {\it scene} reconstruction. 
For instance, the voxel reconstruction of a scene may have no overlap if all the shapes are right, but they are in the wrong place.
Therefore, we formulate it as a 3D detection problem,
as a prediction is a true positive only if all of translation, scale, rotation and shape are correct.
However, 3D detection is a very strict metric.
If the whole scene is slightly off in one aspect, we may have a very low AP. But the predicted scene may still be reasonable.
We mainly use it quantify our performance.

\begin{table}[t]
    \centering
    \scriptsize
    \caption{We report the average precision (AP) in evaluation of the 3D detection setting. \textbf{All} means a prediction is a true positive only if all of translation, scale, rotation and shape are correct. \textbf{Shape}, \textbf{Translation}, \textbf{Rotation}, and \textbf{Scale} mean a prediction is a true positive when a single error is lower than thresholds. We include results on the whole test set, and top 25\%, 50\% and 75\% examples ranked by single-view predictions.}
    \begin{tabular}{c@{\hskip20pt}c@{\hskip5pt}c@{\hskip5pt}c@{\hskip5pt}c@{\hskip5pt}c@{\hskip25pt}c@{\hskip8pt}c@{\hskip8pt}c}
        \toprule
        & \multicolumn{5}{c}{All Examples\quad\quad\quad\quad} &  Top 25\% & Top 50\% & Top 75\%\\
        Methods & All & Shape & Trans & Rot & Scale & All & All & All\\
        \midrule
        Feedforward & 21.2 & 22.5 & 31.7 & 28.5 & 26.9 & 41.6 & 34.6 & 28.6\\
        NMS &  21.1 & 23.5 & 31.9 & 29.0 & 27.2 & 42.0 & 34.7 & 28.7\\
        Raw Affinity & 15.0 & 24.4 & 26.3 & 28.2 & 25.9 & 28.6 & 23.5 & 18.9\\
        \textbf{\papername} & \textbf{23.3} & \textbf{24.5} & \textbf{38.4} & \textbf{29.5} & \textbf{27.3} & \textbf{48.3} & \textbf{38.8} & \textbf{31.4}\\
        \bottomrule
    \end{tabular}
    \label{tab:detection}
\end{table}

Table \ref{tab:detection} shows our performance compared with all three baseline methods.
Our approach outperforms all of them, which verifies what we see in the qualitative examples.
Moreover, the improvement mainly comes from that on translation. The translation-only AP is around 7 points better than Feedforward.
Meanwhile, the improvement of NMS over Feedforward is limited. 
As we see in qualitative examples, it cannot work when many objects are close to each other.
Finally, raw affinity is even worse than Feedforward,
since raw affinity may merge objects incorrectly.
We will discuss why the affinity is informative, but top-1 similarity is not a good choice
in Sec. \ref{sec:expaff}.

We notice our performance gain over Feedforward and NMS is especially large when single-view predictions are reasonable. 
On top 25\% examples which single-view prediction does a good job,
{\papername} outperforms Feedforward and NMS by over 6 points.
On top 50\% examples, the improvement is around 4 points. It is still significant but slightly lower than that of top 25\% examples.
When single-view prediction is bad, our performance gain is limited since {\papername} is built upon it.
We will discuss this in Sec. \ref{sec:expfail} as failure cases.

\subsection{Inter-view Object Affinity Matrix}
\label{sec:expaff}

\begin{table}[t]    
    \centering
    \caption{AUROC and rank correlation between the affinity matrix and {\it category}, {\it model}, {\it shape}, and {\it instance}, respectively. \textbf{Model $|$ Category} means the ability of the affinity matrix to distinguish different models given the same category / semantic label.}
    \begin{tabular}{@{\hskip10pt}l@{\hskip10pt}c@{\hskip10pt}c@{\hskip10pt}c@{\hskip10pt}c@{\hskip10pt}}
    \toprule
    & Category & Model $|$ Category & Shape $|$ Category &  Instance $|$ Model\\
    \midrule
    AUROC & 0.92 & 0.73 & - & 0.59 \\
    Correlation & 0.72 & 0.33 & 0.34 & 0.14 \\
    \bottomrule
    \end{tabular}
    \label{tab:aff_ablation}
\end{table}

We then turn to evaluating how the method works by analyzing
individual components. We start with the affinity matrix and study what it learns. 

We have three non-mutually exclusive hypotheses: 
(1) \textbf{Semantic labels.} The affinity is essentially doing object recognition.
After detecting the category of the object,
it simply matches objects with the same category.
(2) \textbf{Object shapes.} The affinity matches objects with similar shapes since it is constructed from the embedding vectors which are also used to generate shape voxels and the object pose. 
(3) \textbf{Correspondence.} 
Ideally, the affinity matrix should give us ground truth correspondence. 
It is challenging given duplicate objects in the scene. 
For example, people can have three identical chairs in their office. 
These hypotheses are three different levels the affinity matrix may learn, but they are not in conflict. 
Learning semantic labels do not mean the affinity does not learn anything about shapes.

We study this by examining a large number of pairs of objects and testing
the relationship between affinity and known relationships (e.g., categories, model ids) using ground truth bounding boxes.
We specifically
construct three binary labels (same {\it category}, same {\it
model}, same {\it instance}) and a continuous label shape similarity
(namely F-score\,@\,0.05~\cite{tatarchenko2019single}).  When we evaluate shape
similarity, we condition on the category to test if affinity distinguishes
between different models of the same category, (e.g. chair). Similarly, we
condition on the model when we evaluate instance similarity.

We compute two metrics: a binary classification metric that treats the
affinity as a predictor of the label as well as a correlation that tests if a monotonic
relationship exists between the affinity and the label.
For binary classification, we use AUROC to evaluate the performance since it is
invariant to class imbalance and has a natural interpretation.
For correlation, we compute Spearman's rank correlation coefficient~\cite{zwillinger1999crc} between the
affinity predictors and labels. This tests how well the relationship between
affinity and each label (e.g., shape overlap) fits a monotonic function (1 is perfect agreement, 0 no agreement).

The results are shown in Table \ref{tab:aff_ablation}.
Both the binary classification and the rank correlation show that the affinity matrix is able to distinguish different categories and objects of different shapes, but is sub-optimal in distinguishing the same instance.
These results justify our stitching stage, which addresses the problem based on joint reasoning. 
It also explains why Raw Affinity underperforms all other baselines by a large margin in the full-scene evaluation.
Additionally, the ability to distinguish categories and shapes provides important guidance to the stitching stage.
For example, a sofa and bed are
similar in 3D shapes. It is infeasible to distinguish them
by simply looking at the chamfer distance, which can be distinguished by the affinity matrix.

\subsection{Stitching Stage}

We evaluate the stitching stage by studying two questions: (1) How well can it predict object correspondence? (2) Can it improve relative camera pose estimation? For example, if the top-1 relative pose is incorrect, could the stitching stage fix it by considering common objects in two views? 

\begin{figure}[!t]
    \centering
    
    \begin{tabular}{c@{\hskip3pt}c@{\hskip8pt}c@{\hskip3pt}c@{\hskip8pt}c@{\hskip3pt}c}
    \toprule
    
    Before & After & Before & After & Before & After\\
    \midrule
    \includegraphics[width=0.14\textwidth]{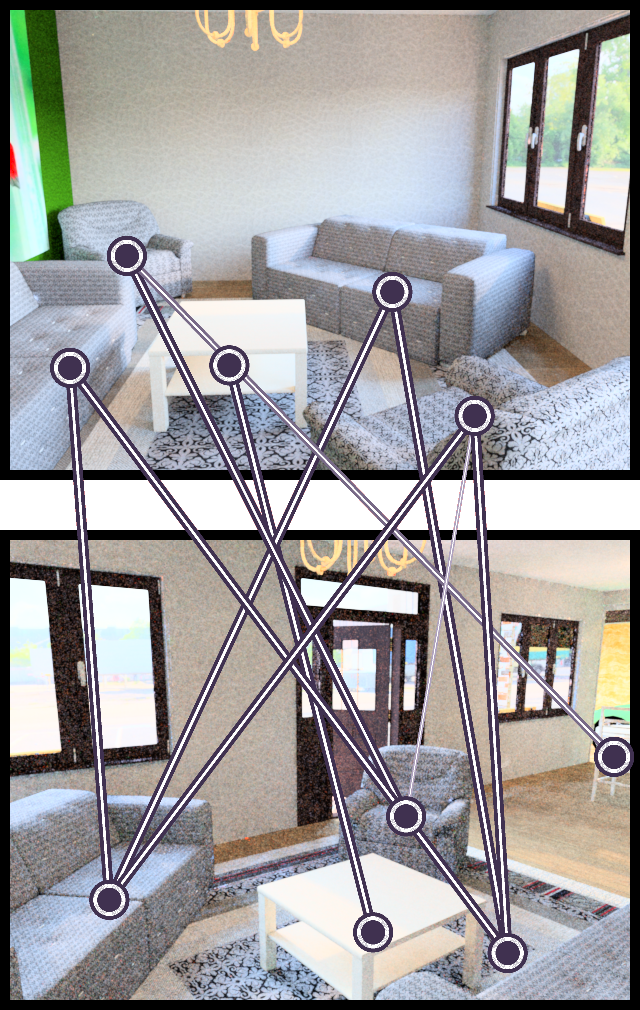}
    & \includegraphics[width=0.14\textwidth]{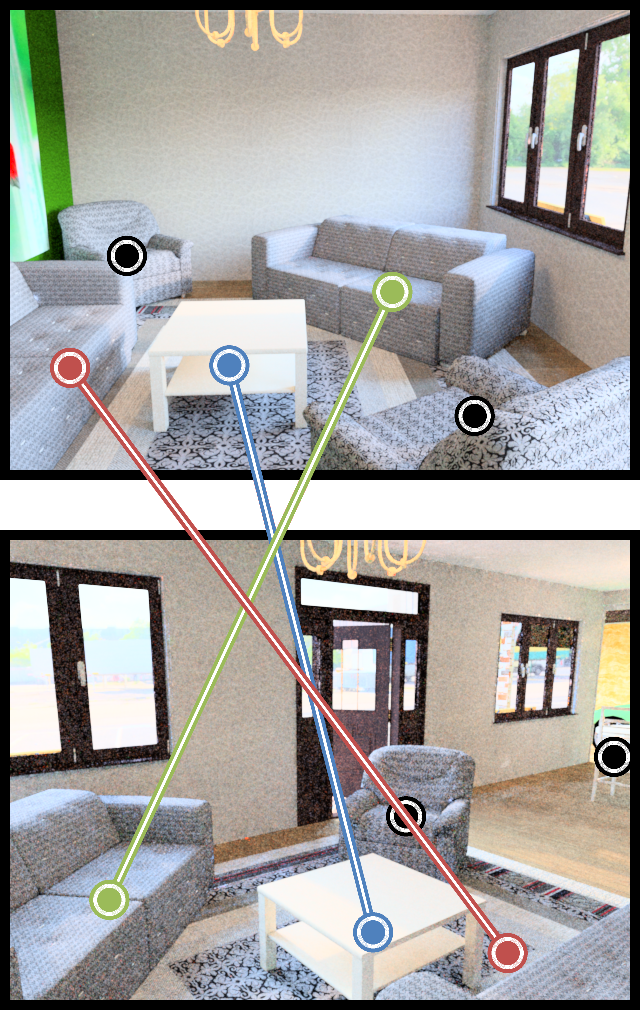}
    & \includegraphics[width=0.14\textwidth]{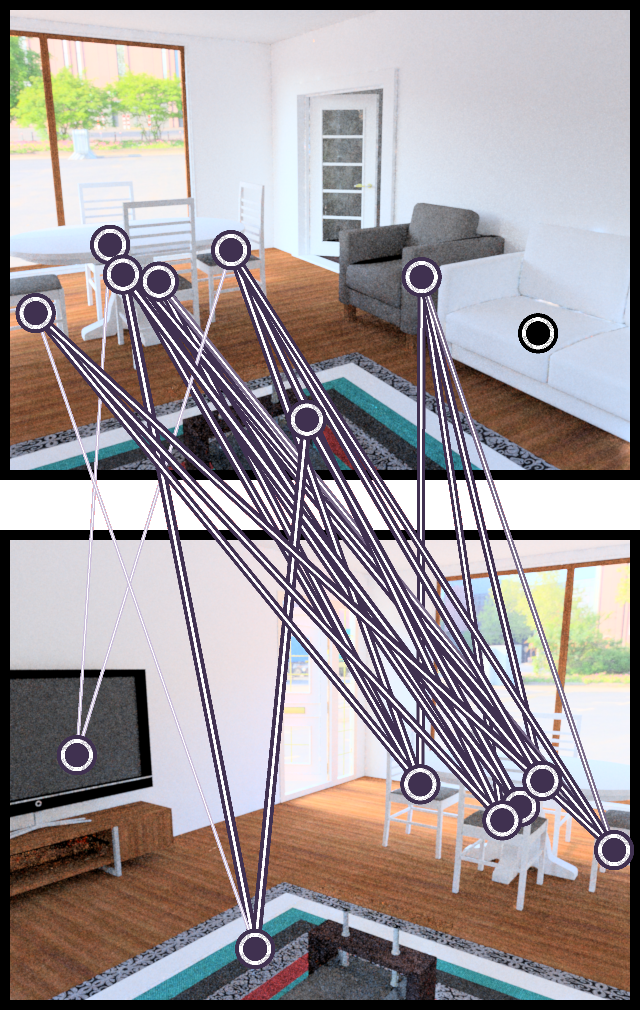}
    & \includegraphics[width=0.14\textwidth]{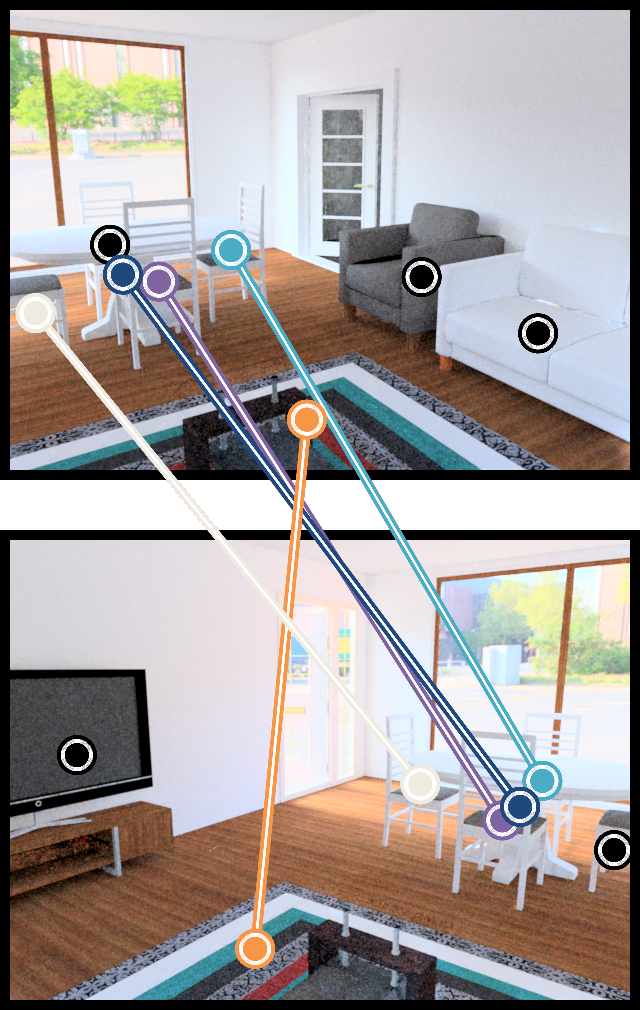}
    & \includegraphics[width=0.14\textwidth]{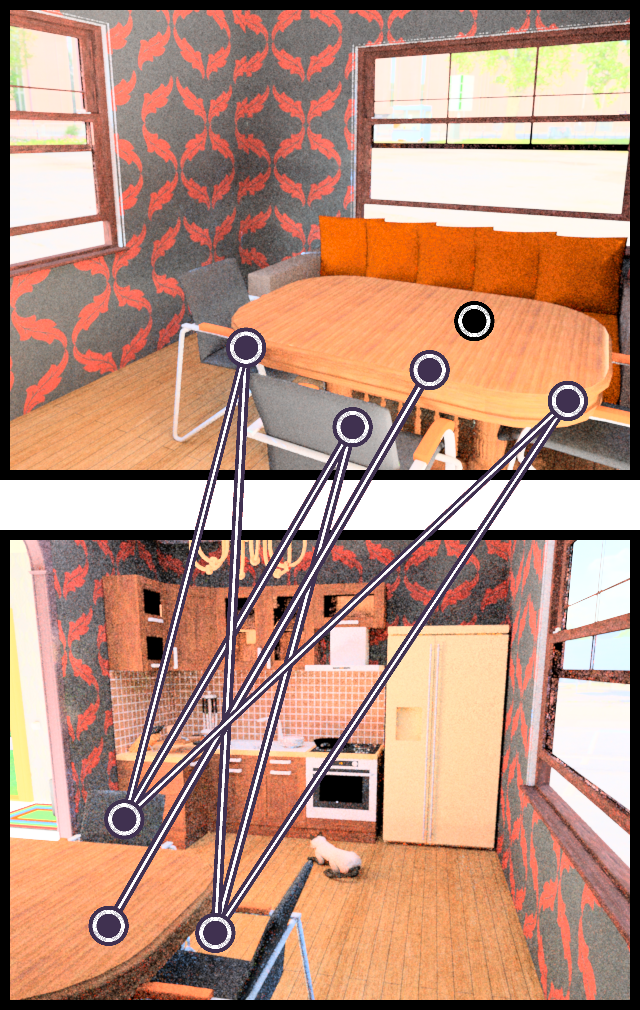}
    & \includegraphics[width=0.14\textwidth]{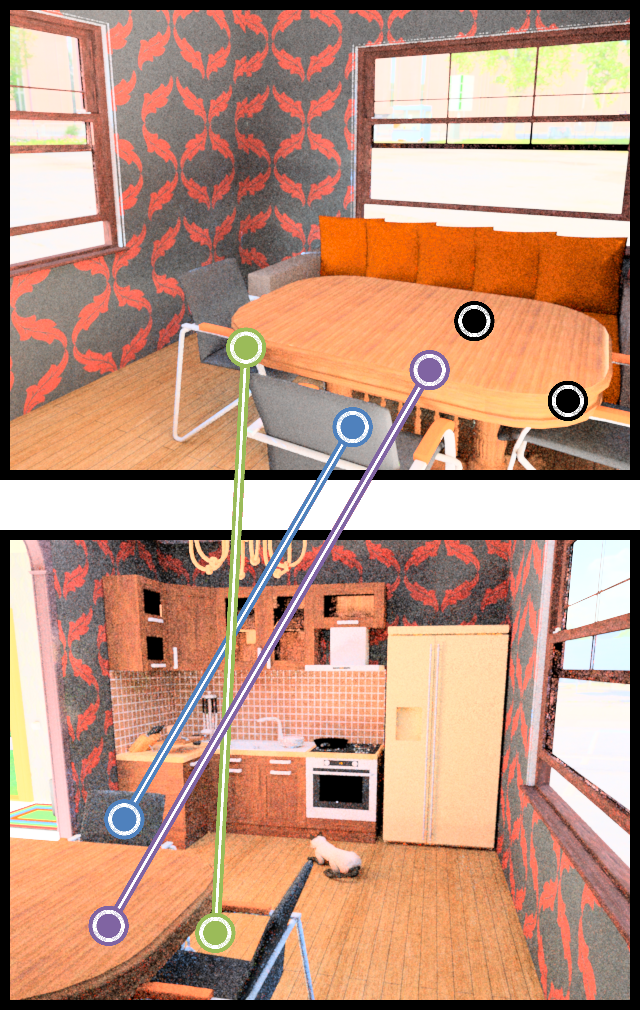}\\

    \bottomrule
    \end{tabular}
    \caption{Visualization of the stitching stage. The affinity matrix generates proposals of corresponding objects, and then the stitching stage removes outliers by inferring the most likely explanation of the scene.}
    \label{fig:stitching}
\end{figure}

\vspace{0.75em}
\noindent
\textbf{Object Correspondence.} 
To answer the first question, we begin with qualitative examples in Fig. \ref{fig:stitching}, which illustrate object correspondence before and after the stitching stage. 
Before our stitching stage, our affinity matrix has generated correspondence proposals based on their similarity. 
However, there are outliers since the affinity is sub-optimal in distinguishing the same instance. 
The stitching stage removes these outliers.

We evaluate object correspondence in the same setting as Sec \ref{sec:expaff}. 
Suppose the first and second images have $N$ and $M$ objects respectively. We then have $N \times M$ pairs. 
The pair is a positive example if and only if they are corresponding.
We use average precision (AP) to measure the performance since AP pays more attention to the low recall~\cite{davis2006relationship,everingham2010pascal}.
For $i^{th}$ object in view 1 and $j^{th}$ object in view 2, we produce a confidence score by $\gamma A_{ij}$ where $\gamma = 1$ if the pair is predicted to be corresponding and $\gamma = 0.5$ otherwise. 
This $\gamma$ term updates the confidence based on stitching stage to penalize pairs which have a high affinity score but are not corresponding.

We compare {\papername} with 3 baselines. 
(\textbf{All Negative}): The prediction is always negative (the most frequent label). 
This serves as a lower bound. 
(\textbf{Affinity}): This simply uses the affinity matrix as the confidence. 
(\textbf{Affinity Top1}): Rather than using the raw affinity matrix, 
it uses affinity top-1 similarity as the correspondence and the same strategy to decide confidence as {\papername}.
Table~\ref{tab:correspondence} shows that our stitching stage improves AP by 10\% compared to using the affinity matrix only as correspondence.

\begin{table}[t]
    \centering
    \caption{Evaluation of object correspondence with and without the stitching stage.}
    \begin{tabular}{@{\hskip10pt}c@{\hskip20pt}c@{\hskip20pt}c@{\hskip20pt}c@{\hskip20pt}c@{\hskip10pt}}
        \toprule
        & All Negative & Affinity & Affinity Top1 & \textbf{\papername}\\
        \midrule
        AP & 10.1 & 38.8 & 49.4 & \textbf{60.0}\\
        \bottomrule
    \end{tabular}
    \label{tab:correspondence}
\end{table}

\begin{table}[t]    
    \centering
    \scriptsize
    \caption{Evaluation of relative camera pose from the camera branch and picked by the stitching.}
    \begin{tabular}{@{\hskip8pt}c@{\hskip14pt}c@{\hskip9pt}c@{\hskip9pt}c@{\hskip14pt}c@{\hskip9pt}c@{\hskip9pt}c@{\hskip8pt}}
    \toprule
    & \multicolumn{3}{c}{Translation (meters)} & \multicolumn{3}{c}{Rotation  (degrees)} \\
    Method & Median & Mean & (Err $\leq$ 1m)\% & Median    & Mean & (Err $\leq$ 30$^{ \circ }$)\% \\
    \midrule
    Top-1   & 1.24     & 1.80 & 41.26 & \textbf{6.96} & 29.90 & 77.56 \\
    \textbf{\papername} & \textbf{0.88} & \textbf{1.44} & \textbf{54.89} & 6.97 & \textbf{29.02} & \textbf{78.31} \\
    \bottomrule
    \end{tabular}
    \label{tab:relpose}
\end{table}

\vspace{0.75em}
\noindent
\textbf{Relative Camera Pose Estimation.} We next evaluate the performance of relative camera pose (i.e., camera translation and rotation) estimation and see if the stitching stage improves the relative camera pose jointly.
We compare the camera pose picked by the stitching stage and top-1 camera pose predicted by the camera branch.
We follow the rotation and translation metrics in our full-scene evaluation to measure the error of our predicted camera poses.
We summarize results in Table \ref{tab:relpose}.
There is a substantial improvement in translations, with the percentage of camera poses 
within $1$m of the ground truth being boosted from $41.3\%$ to $54.9\%$. 
The improvement in rotation is smaller and we believe
this is because the network already starts out working well and can exploit the fact that scenes tend to have three orthogonal directions. In conclusion, the stitching stage can mainly improve the prediction of camera translation.

\begin{figure*}[t]
    \centering
    
    \begin{tabular}{c@{\hskip3pt}cc@{\hskip3pt}cc@{\hskip3pt}cc@{\hskip3pt}c}
        \toprule
    
    \multicolumn{2}{c}{Input Images} & \multicolumn{2}{c}{Camera 1} &\multicolumn{2}{c}{Camera 2} & \multicolumn{2}{c}{Birdview} \\
    Image 1 & Image 2 & Prediction & GT & Prediction & GT & Prediction & GT\\
    \midrule
    \multicolumn{2}{c}{\includegraphics[width=0.23\textwidth]{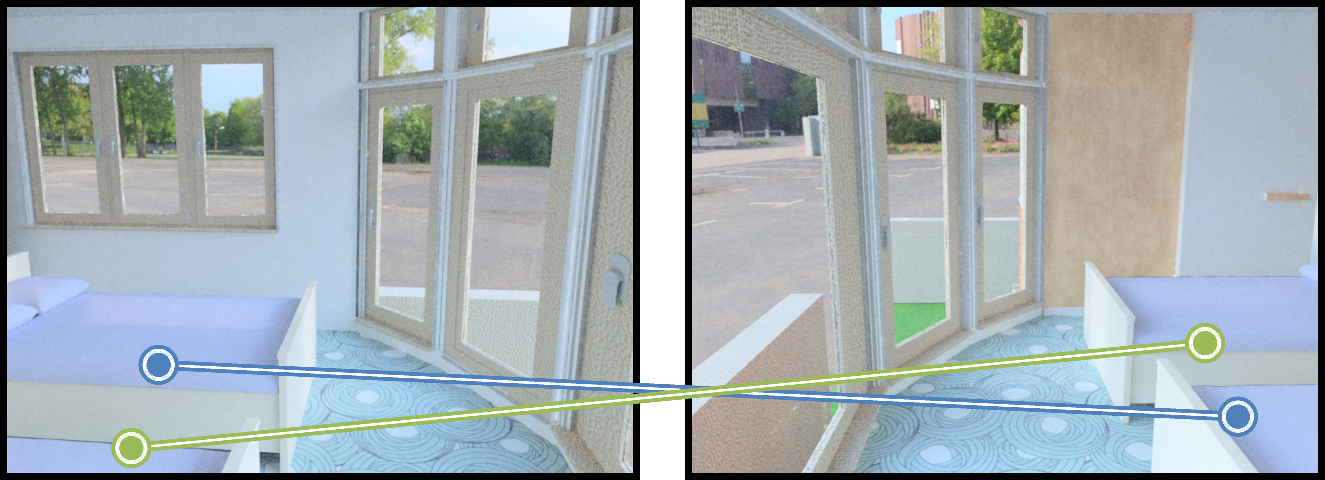}}
    & \frame{\includegraphics[width=0.11\textwidth]{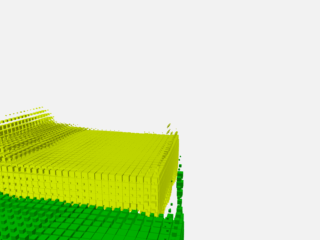}}
    & \frame{\includegraphics[width=0.11\textwidth]{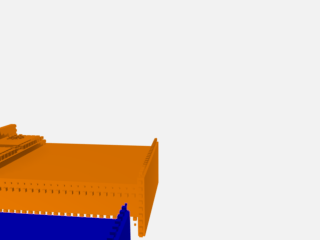}}
    & \frame{\includegraphics[width=0.11\textwidth]{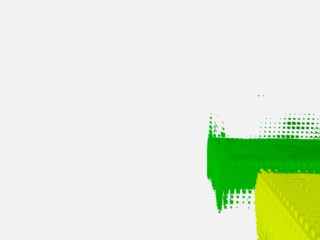}}
    & \frame{\includegraphics[width=0.11\textwidth]{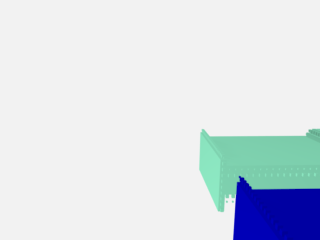}}
    & \frame{\includegraphics[width=0.11\textwidth]{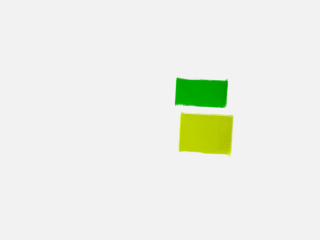}}
    & \frame{\includegraphics[width=0.11\textwidth]{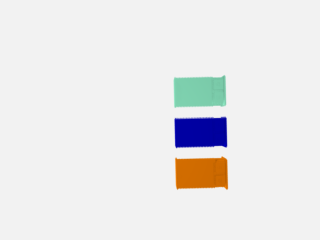}}\\

    \multicolumn{2}{c}{\includegraphics[width=0.23\textwidth]{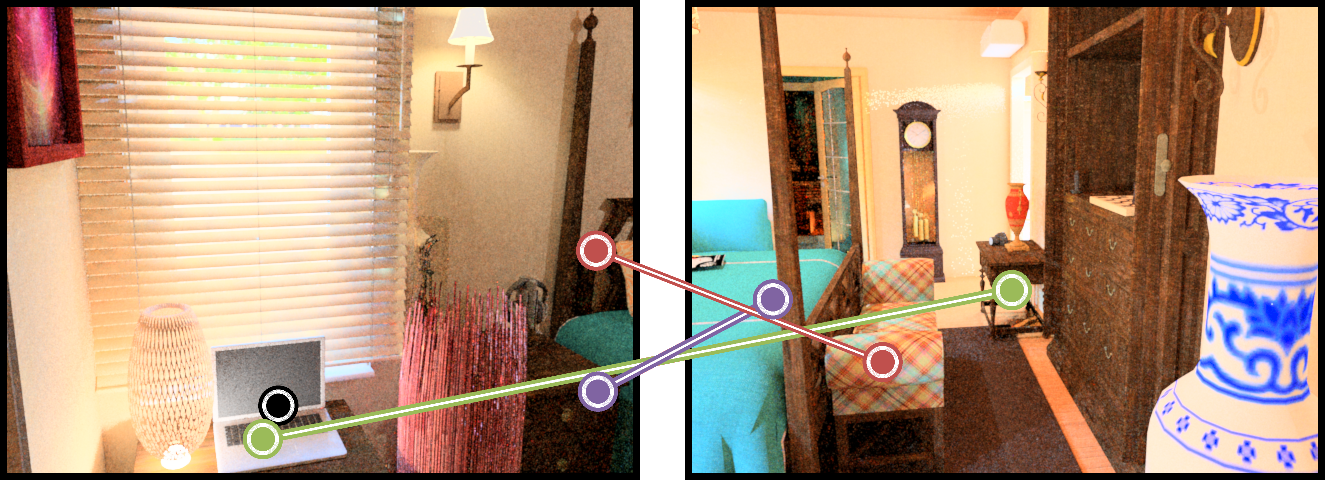}}
    & \frame{\includegraphics[width=0.11\textwidth]{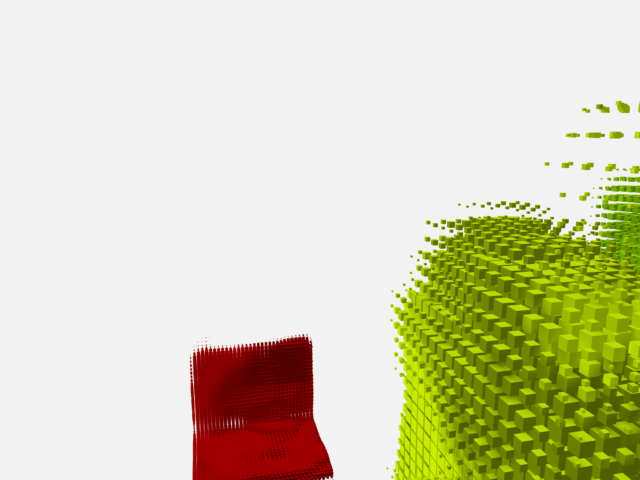}}
    & \frame{\includegraphics[width=0.11\textwidth]{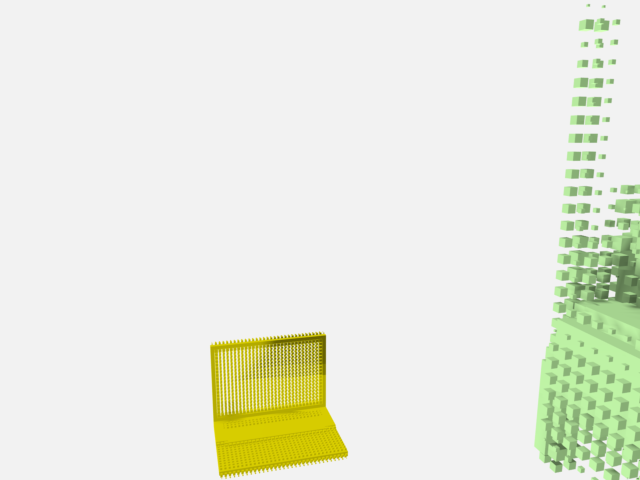}}
    & \frame{\includegraphics[width=0.11\textwidth]{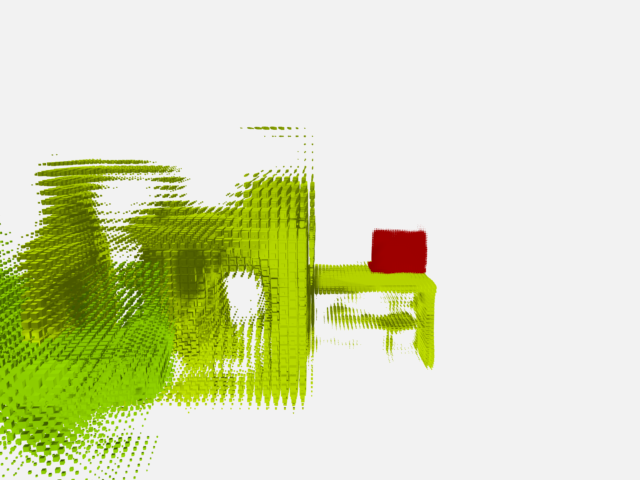}}
    & \frame{\includegraphics[width=0.11\textwidth]{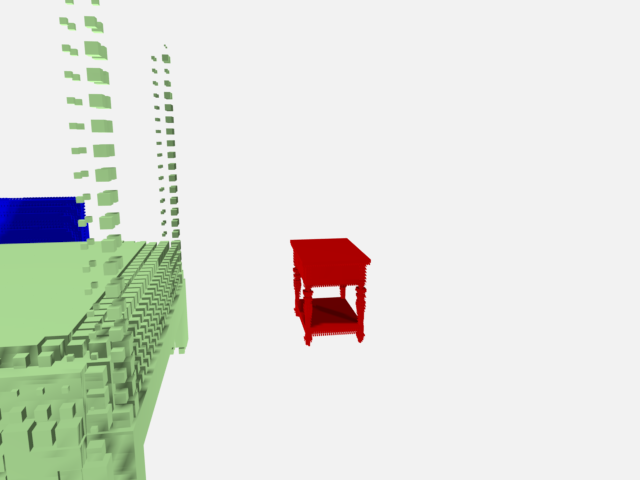}}
    & \frame{\includegraphics[width=0.11\textwidth]{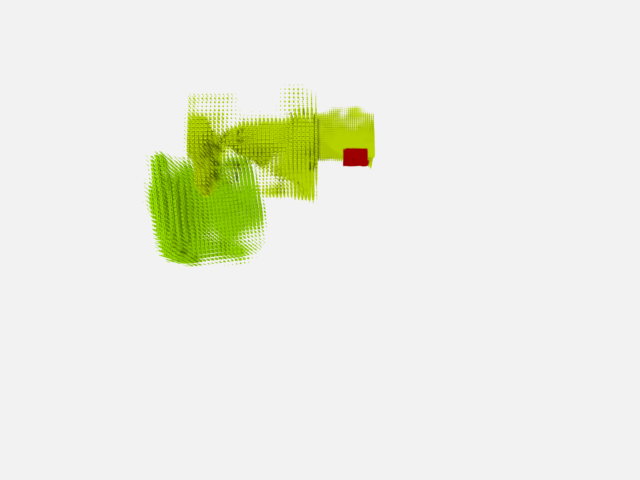}}
    & \frame{\includegraphics[width=0.11\textwidth]{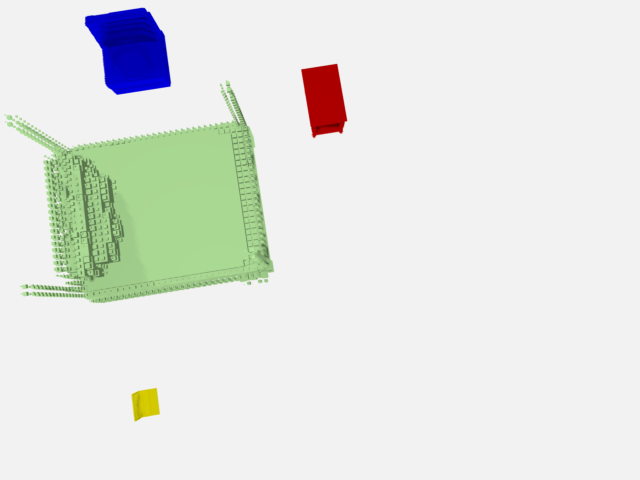}}\\

    \multicolumn{2}{c}{\includegraphics[width=0.23\textwidth]{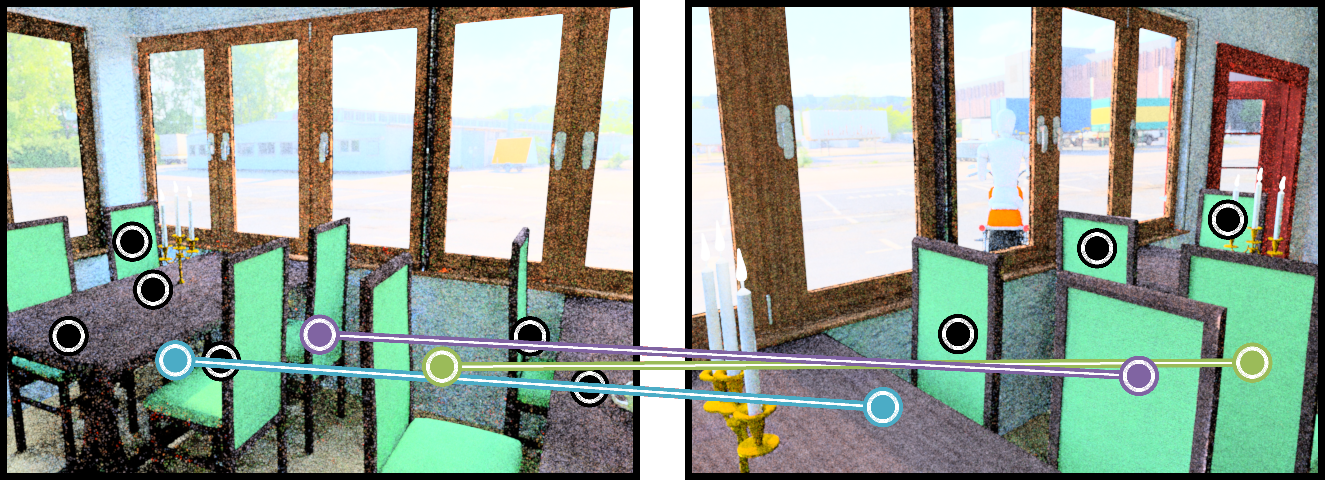}}
    & \frame{\includegraphics[width=0.11\textwidth]{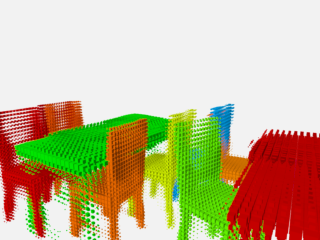}}
    & \frame{\includegraphics[width=0.11\textwidth]{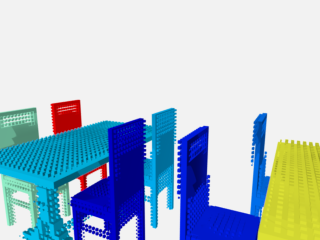}}
    & \frame{\includegraphics[width=0.11\textwidth]{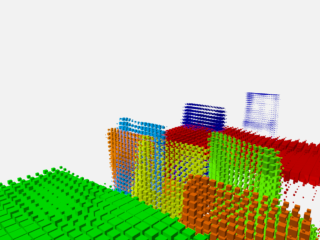}}
    & \frame{\includegraphics[width=0.11\textwidth]{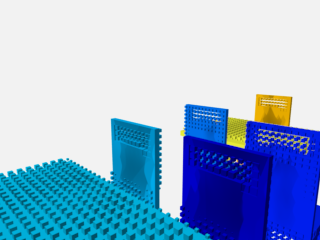}}
    & \frame{\includegraphics[width=0.11\textwidth]{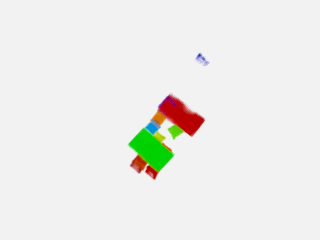}}
    & \frame{\includegraphics[width=0.11\textwidth]{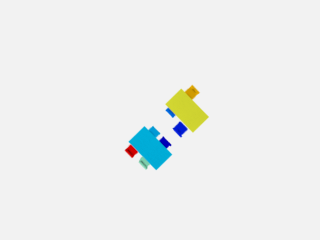}}\\
    \bottomrule
    \end{tabular}
    \caption{Representative failure cases on the SUNCG test set \cite{song2017semantic}. \textbf{Row 1:} The input images are ambiguous. There can be two or three beds in the scene. \textbf{Row 2:} The single-view backbone does not produce a reasonable prediction. \textbf{Row 3:} This is challenging because all chairs are the same.}
    \label{fig:suncg-failure}
\end{figure*}

\subsection{Failure Cases}
\label{sec:expfail}

To understand the problem of reconstruction from sparse views better, 
we identify some representative failure cases 
and show them in Fig. \ref{fig:suncg-failure}.
While our method is able to generate reasonable 
results on SUNCG, it cannot solve some common failure cases:
(1) The image pair is ambiguous.
(2) The single-view backbone does not produce reasonable predictions as we discuss in Sec. \ref{sec:expfull}.
(3) There are too many similar objects in the scene. 
The affinity matrix is then not able to distinguish them since
it is sub-optimal in distinguishing the same instance.
Our stitching 
stage is also limited by the random search
over object correspondence. 
Due to factorial growth of search space, 
we cannot search all possible correspondences.
The balancing of our sub-losses can also be sensitive.

\begin{figure*}[!t]
    \centering
    \begin{tabular}{@{\hskip3pt}c@{\hskip3pt}c@{\hskip5pt}c@{\hskip3pt}c@{\hskip10pt}c@{\hskip3pt}c@{\hskip5pt}c@{\hskip3pt}c}
        \toprule
        Image 1 & Image 2 & Sideview & Birdview & Image 1 & Image 2 & Sideview & Birdview\\
        \midrule
        \frame{\includegraphics[width=0.11\textwidth]{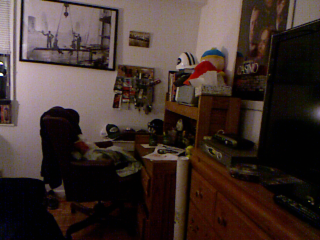}} & 
        \frame{\includegraphics[width=0.11\textwidth]{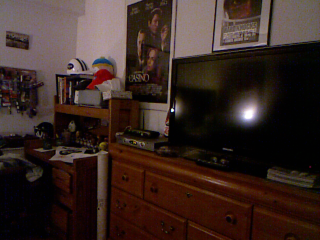}} & 
        \frame{\includegraphics[width=0.11\textwidth]{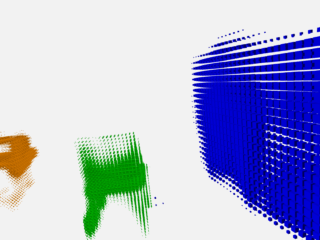}} & 
        \frame{\includegraphics[width=0.11\textwidth]{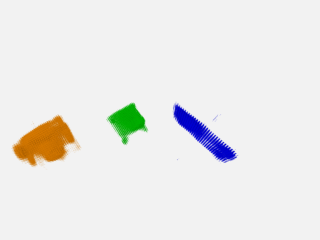}} &

        \frame{\includegraphics[width=0.11\textwidth]{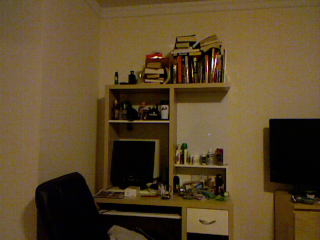}} & 
    \frame{\includegraphics[width=0.11\textwidth]{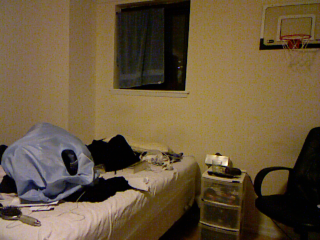}} & 
    \frame{\includegraphics[width=0.11\textwidth]{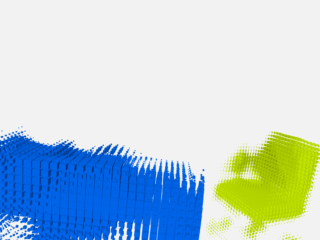}} & 
    \frame{\includegraphics[width=0.11\textwidth]{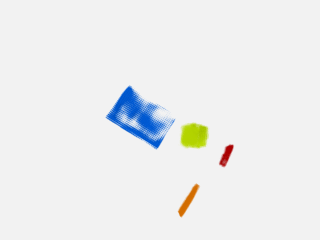}} \\

    \frame{\includegraphics[width=0.11\textwidth]{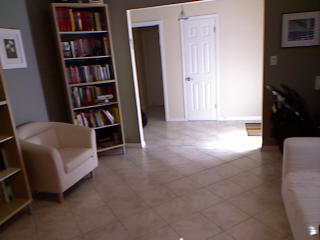}} & 
    \frame{\includegraphics[width=0.11\textwidth]{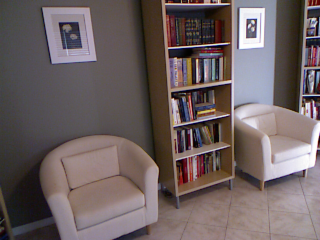}} & 
    \frame{\includegraphics[width=0.11\textwidth]{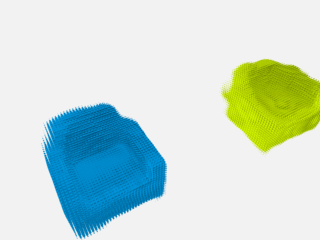}} & 
    \frame{\includegraphics[width=0.11\textwidth]{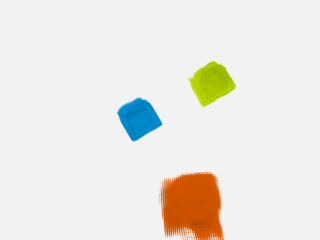}} &
    
    \frame{\includegraphics[width=0.11\textwidth]{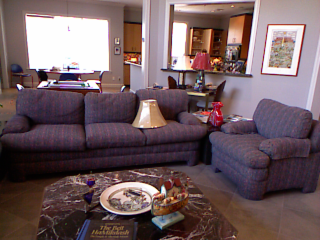}} & 
    \frame{\includegraphics[width=0.11\textwidth]{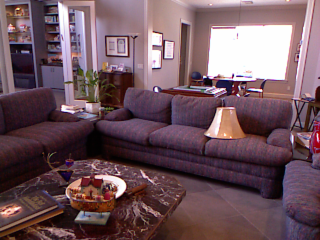}} & 
    \frame{\includegraphics[width=0.11\textwidth]{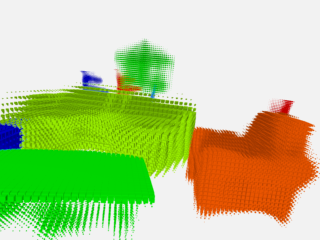}} & 
    \frame{\includegraphics[width=0.11\textwidth]{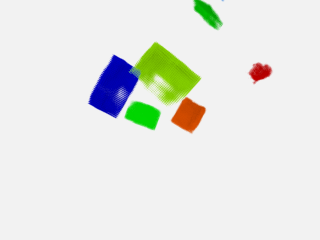}} \\

        \bottomrule
    \end{tabular}
    \caption{Qualitative results on NYUv2 dataset \cite{Silberman12}. Sideview corresponds to the camera view slightly transformed from the image 2 camera position.}
    \label{fig:nyu-wall}
\end{figure*}

\subsection{Results on NYU Dataset}
\label{sec:expnyu}
To test generalization, we also test our approach on images from NYUv2
\cite{Silberman12}. 
Our only change is using proposals
from Faster-RCNN \cite{ren2015faster} trained on COCO \cite{lin2014microsoft}, since Faster-RCNN trained on SUNCG cannot generalize to NYUv2 well. 
We do not finetune any models and show qualitative results in Fig. \ref{fig:nyu-wall}. 
Despite training on synthetic data, our model can 
often obtain a reasonable interpretation.

\section{Conclusion}
\label{sec:conclusion}
We have presented {\papername}, which explores 3D volumetric reconstruction
from sparse views.
While the output is reasonable, failure modes indicate the problem is challenging to current techniques.
Directions for future work include joint learning of object affinity and relative camera pose, and extending the approach to many views and more natural datasets other
than SUNCG.

\vspace{0.75em}
\noindent
{\bf Acknowledgments} 
We thank Nilesh Kulkarni and Shubham Tulsiani for their help of 3D-RelNet; 
Zhengyuan Dong for his help of visualization;
Tianning Zhu for his help of video;
Richard Higgins, Dandan Shan, Chris Rockwell and Tongan Cai for their feedback on the draft. Toyota Research Institute (``TRI'') provided funds to assist the authors with their research but this article solely reflects the opinions and conclusions of its authors and not TRI or any other Toyota entity.

\bibliographystyle{splncs04}
\bibliography{local}

\clearpage
\appendix
\section{Implementation}

\noindent
\textbf{Detection proposals.} 
We use more advanced object proposals compared to prior works \cite{kulkarni20193d,tulsiani2018factoring},
which used edge boxes \cite{zitnick2014edge}. We found that edge boxes
were often the limiting factor. 
Instead, we train a class-agnostic Faster-RCNN \cite{ren2015faster} to generate proposals, treating all objects as the foreground.

\vspace{0.75em}
\noindent
\textbf{Object Branch.} 
For each object, our object branch will predict a 300-dimensional vector, which represents its 3D properties. 
Linear layers are used to predict its shape embedding, translation, rotation, scale and object embedding. 
For the object embeddings, we use three linear layers. 
The size of outputs is 256, 128, 64, respectively. 
These linear layers predict a 64-dimensional embedding finally.

We train the object branch in two stages.
In the first stage, we follow the training of 3D-RelNet \cite{kulkarni20193d}
with ground truth bounding boxes. 
The loss of affinity matrix is ignored in this stage. 
In the second stage, we freeze all layers except the linear layers to predict the object embeddings.
We only apply the affinity loss in this stage. 
For all two stages, we use Adam with learning rate $\varepsilon = 10^{-4}$ to optimize the model, with momentum 0.9. The batch size is 24.
Although 3D-RelNet is finetuned on detection proposals, 
we only use the intermediate model trained with ground truth bounding box 
because (1) 3D-RelNet is finetuned on edgebox proposals and our Faster-RCNN proposals are good enough; (2) ground truth affinity is only available with ground truth bounding box.

\vspace{0.75em}
\noindent
\textbf{Camera Branch.}
The object and camera branches are trained independently.
The translation is represented as 3D vectors, and the rotation is represented as quaternions.
We run k-means clustering on the training set to produce 60 and 30 bins for translation and rotation. 
For rotation, we use spherical k-means to ensure the centroids are unit vectors.

The input image pairs are resized to 224x224. 
They are passed through a siamese network with ResNet-50 pre-trained on ImageNet \cite{he2016deep} as the backbone.
The outputs from each instance of the siamese network are concatenated, and passed through a linear layer, producing a 128-dimensional vector.
The vector is then passed through a translation branch and a rotation branch. 
Each branch is a linear layer which outputs 60 and 30 dimensional vectors for translation and rotation bins.

Our loss function is the cross entropy loss. 
The loss for the translation prediction and the rotation prediction are weighted equally.
We use stochastic gradient descent with learning rate $\varepsilon = 10^{-3}$ and momentum $0.9$.
The batch size is 32. 
We also augment the data by reversing the order of image pairs.

\vspace{0.75em}
\noindent
\textbf{Tuning the stitching stage.} 
The search space contains 
top-3 rotation, top-10 translation and top-128 object 
correspondence hypotheses. The threshold of affinity is 0.5. $\lambda_P$, $\lambda_S$, 
and $\lambda_U$ are tuned as hyperparameters on the 
validation set to preclude the trivial solution. 
We use $\lambda_S = 5$, $\lambda_U = 1$.
For $\lambda_P$, we use 5 for rotation and 1 for translation.

\section{Visualization of the Object Embedding Space}
\begin{figure}[h]
    \centering
    \includegraphics[width=0.45\textwidth]{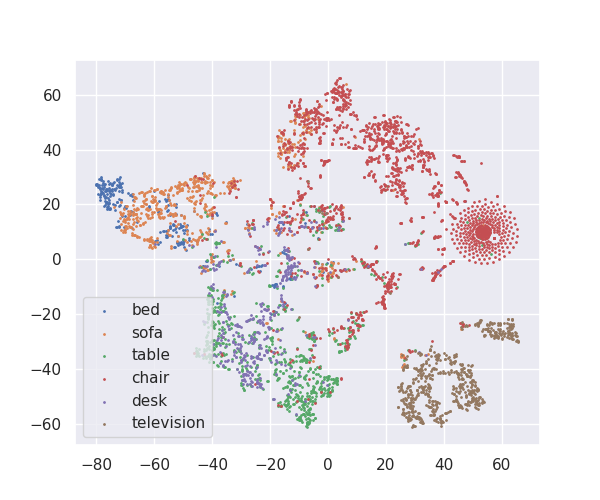}
    \includegraphics[width=0.45\textwidth]{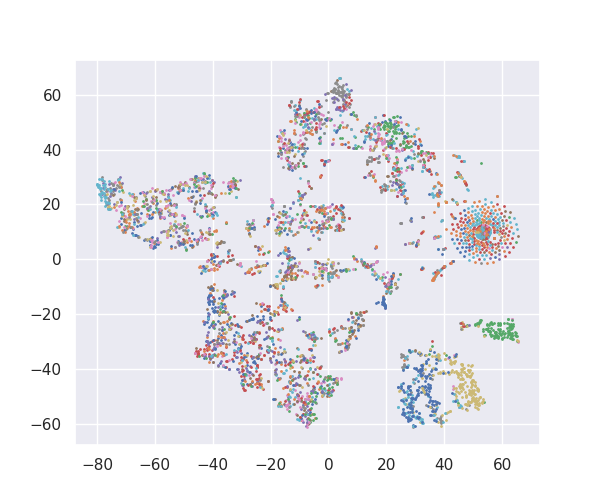}
    \caption{t-SNE visualization of the object embedding space. \textbf{Left:} We assign the same color to embeddings with the same semantic labels. \textbf{Right:} We assign the same color to embeddings with the same 3D models.}
    \label{fig:tsne}
\end{figure}

We use t-SNE~\cite{maaten2008visualizing} to visualize the object embedding space to check what the affinity matrix learns visually. We show our results in Fig.~\ref{fig:tsne}. 

Without using semantic labels as supervision, objects in the same category are closer to each other. 
We also notice that table and desk have similar embeddings.
The object embeddings can distinguish 3D models partially, but not as well as semantic labels.

\begin{figure}[h]
    \centering
    \includegraphics[width=0.45\textwidth]{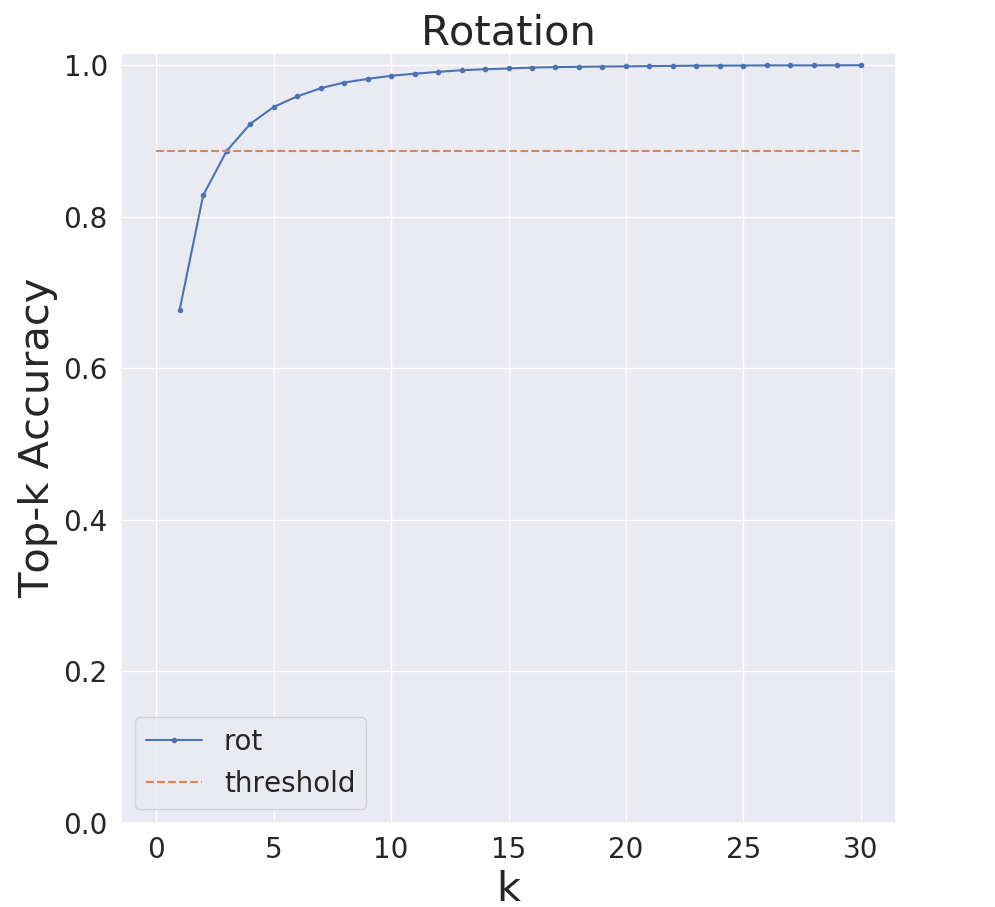}
    \includegraphics[width=0.45\textwidth]{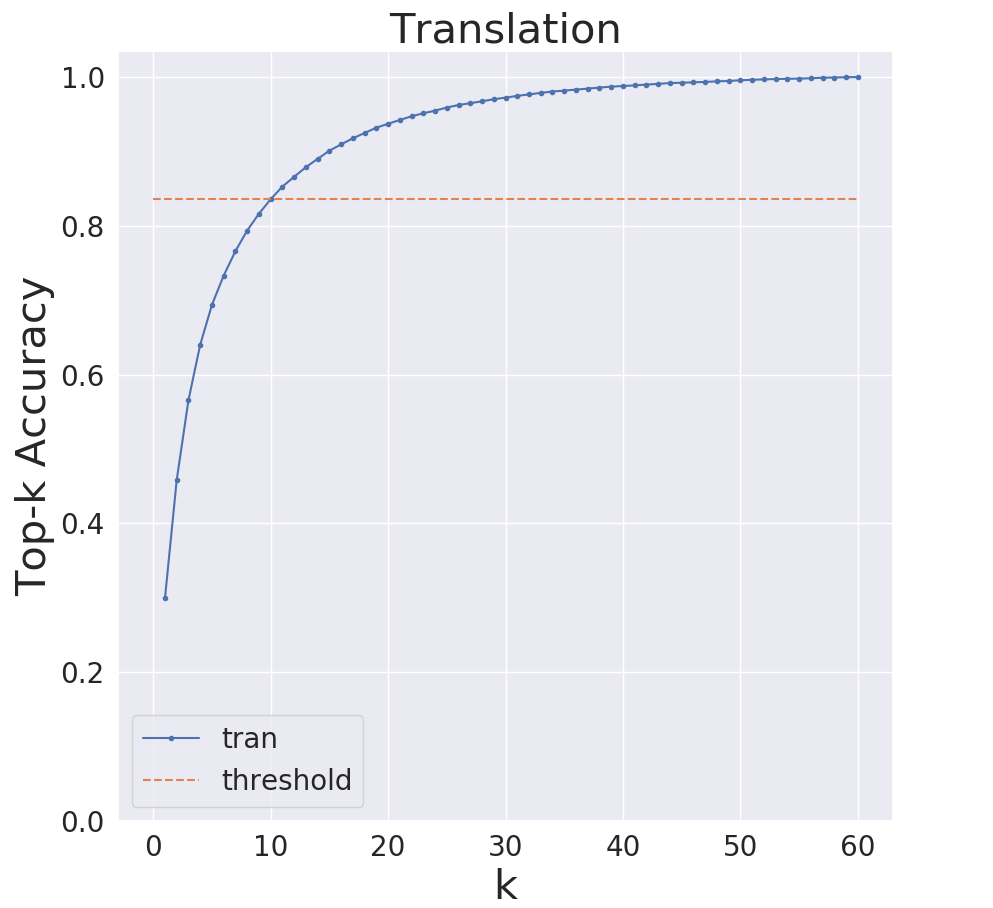}
    \caption{Top-K accuracy of the camera branch on the validation set. The orange line shows the Top-K accuracy of K we choose for the stitching stage.}
    \label{fig:topk_eval}
\end{figure}

\section{Proposals of Camera Pose Transformation}

In the stitching stage, we select the top 3 most likely bins for rotation and top 10 most likely bins for translation.
We demonstrate our motivation for choosing the number.
On the validation set, we evaluate the top-K classification accuracy.
We show the top-K accuracy for translation and rotation in Fig.~\ref{fig:topk_eval}.
We notice that the top-1 accuracy is not high.
However, the top 3 most likely bins for rotation and top 10 most likely bins for translation can ensure a high accuracy in a relatively small searching space.
Therefore, we select these bins in the stitching stage.
During test time, the top-3 rotation accuracy is 88.7\% and top-10 translation accuracy is 83.6\%.

\section{Comparion with Single-view Baselines}

In full scene evaluation, we are also interested in 
whether the multi-view setting helps, since we only add another sparse view.
We address the question by comparing with single-view baselines. 
We take a prediction from 3D-RelNet~\cite{kulkarni20193d} on one view of the pair, selected randomly. On the whole test set, the AP is 13.7, which significantly underperforms all baselines. 
It shows our proposed approach has significant improvement built upon single-view baselines, and multi-view helps reconstruct the scene.

\section{Merging Corresponding Objects}

When our approach finds corresponding objects in two views,
we average the translation and scale, but pick up the shape and rotation at random.
Here we study alternative options.
We use top 50\% examples in the test set ranked by the performance of single-view predictions,
so that the difference is more obvious.

First, we empirically show the rotation cannot be averaged, since there are typically multiple rotation modes. 
In Table~\ref{tab:rot}, we compare the peformance between averaging rotation and pick up one at random.
Averaging rotation dramatically hurts the performance.

\begin{table}[h]
    \centering
    \caption{Comparison between averaging the rotation and picking up one at random.}
    \begin{tabular}{c@{\hskip30pt}c@{\hskip25pt}c@{\hskip25pt}c@{\hskip25pt}c@{\hskip25pt}c@{\hskip10pt}}
        \toprule
         & All & Shape & Trans & Rot & Scale \\
        \midrule
        random rot & 38.8 & 27.3 & 39.6 & 33.2 & 35.1 \\
        average rot & 31.4 & 27.3 & 40.2 & 29.2 & 35.1\\
        \midrule
        improvement & -7.4 & +0.0 & +0.6 & -4.0 & +0.0 \\
        \bottomrule
    \end{tabular}
    \label{tab:rot}
\end{table}

Moreover, a reasonable way to average the shape is to average the vector representation of the object~\cite{Girdhar16b}. 
In Table~\ref{tab:shape}, we compare their performance and they are almost the same.
Therefore, we choose the simplest approach, picking up one shape at random.

\begin{table}[h]
    \centering
    \caption{Comparison between averaging the shape and picking up one at random.}
    \begin{tabular}{c@{\hskip30pt}c@{\hskip25pt}c@{\hskip25pt}c@{\hskip25pt}c@{\hskip25pt}c@{\hskip10pt}}
        \toprule
         & All & Shape & Trans & Rot & Scale \\
        \midrule
        average shape & 38.8 & 27.3 & 39.6 & 33.2 & 35.1 \\
        random shape & 38.8 & 27.2 & 39.6 & 33.2 & 35.1\\
        \midrule
        improvement & +0.0 & -0.1 & +0.0 & +0.0 & +0.0 \\
        \bottomrule
    \end{tabular}
    \label{tab:shape}
\end{table}

\section{Additional Qualitative Results}

\begin{figure*}[t]
    \centering
    
    \begin{tabular}{cccccccc}
        \toprule
    
    \multicolumn{2}{c}{Input Images} & \multicolumn{2}{c}{Camera 1} &\multicolumn{2}{c}{Camera 2} & \multicolumn{2}{c}{Birdview} \\
    Image 1 & Image 2 & Prediction & GT & Prediction & GT & Prediction & GT\\
    \midrule
    \frame{\includegraphics[width=0.12\textwidth]{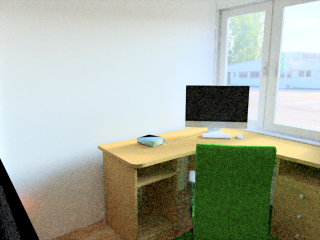}}
    & \frame{\includegraphics[width=0.12\textwidth]{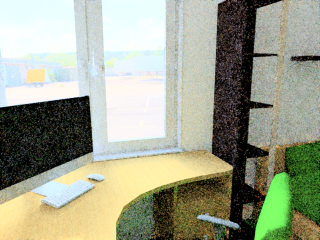}}
    & \frame{\includegraphics[width=0.12\textwidth]{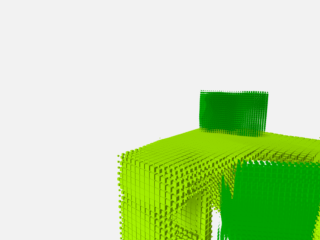}}
    & \frame{\includegraphics[width=0.12\textwidth]{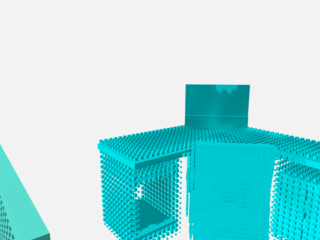}}
    & \frame{\includegraphics[width=0.12\textwidth]{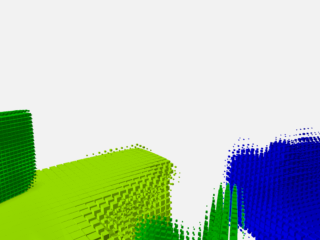}}
    & \frame{\includegraphics[width=0.12\textwidth]{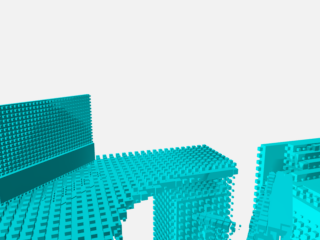}}
    & \frame{\includegraphics[width=0.12\textwidth]{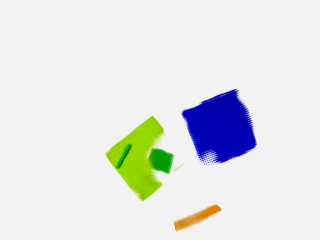}}
    & \frame{\includegraphics[width=0.12\textwidth]{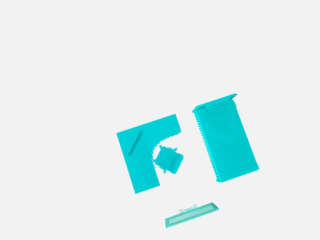}}\\
    
    \frame{\includegraphics[width=0.12\textwidth]{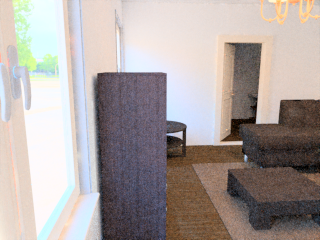}}
    & \frame{\includegraphics[width=0.12\textwidth]{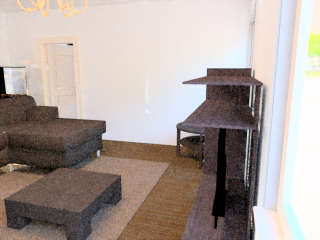}}
    & \frame{\includegraphics[width=0.12\textwidth]{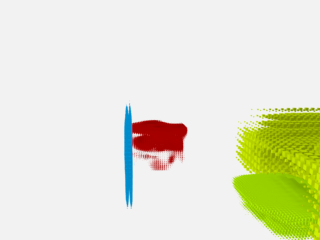}}
    & \frame{\includegraphics[width=0.12\textwidth]{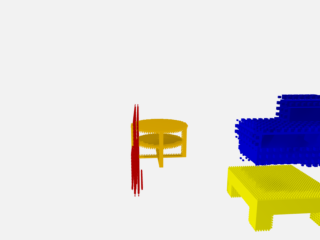}}
    & \frame{\includegraphics[width=0.12\textwidth]{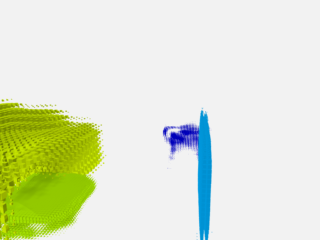}}
    & \frame{\includegraphics[width=0.12\textwidth]{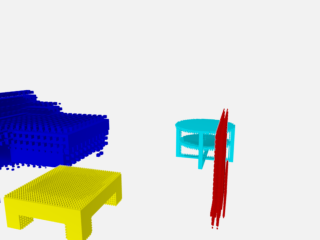}}
    & \frame{\includegraphics[width=0.12\textwidth]{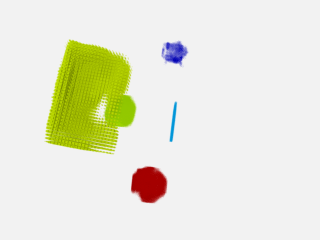}}
    & \frame{\includegraphics[width=0.12\textwidth]{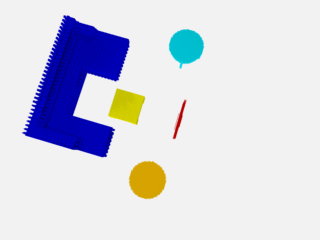}}\\
    
    \frame{\includegraphics[width=0.12\textwidth]{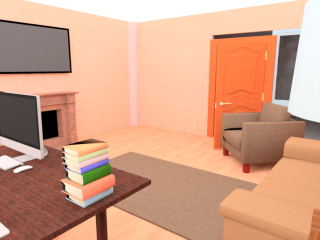}}
    & \frame{\includegraphics[width=0.12\textwidth]{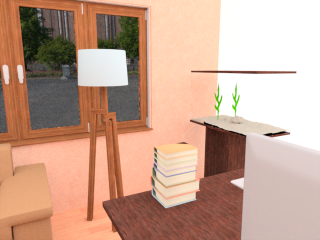}}
    & \frame{\includegraphics[width=0.12\textwidth]{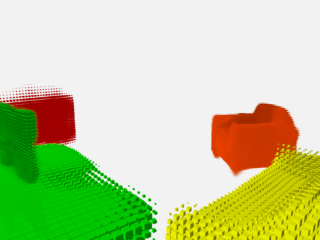}}
    & \frame{\includegraphics[width=0.12\textwidth]{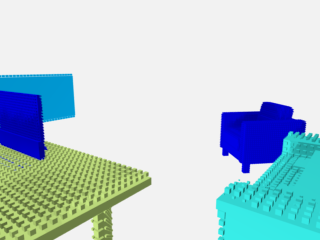}}
    & \frame{\includegraphics[width=0.12\textwidth]{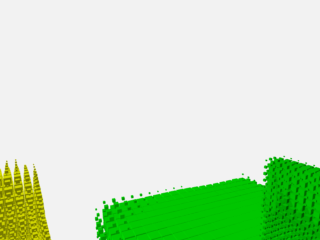}}
    & \frame{\includegraphics[width=0.12\textwidth]{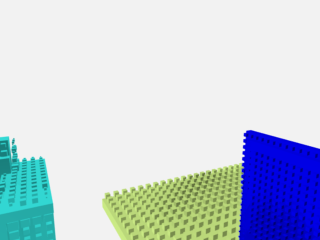}}
    & \frame{\includegraphics[width=0.12\textwidth]{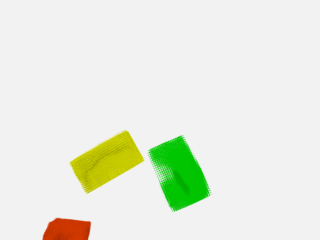}}
    & \frame{\includegraphics[width=0.12\textwidth]{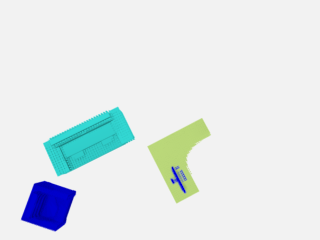}}\\
    
    \frame{\includegraphics[width=0.12\textwidth]{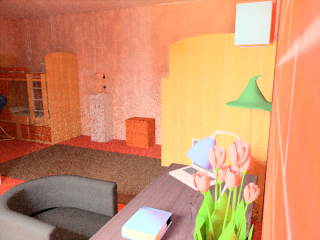}}
    & \frame{\includegraphics[width=0.12\textwidth]{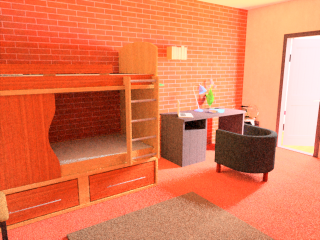}}
    & \frame{\includegraphics[width=0.12\textwidth]{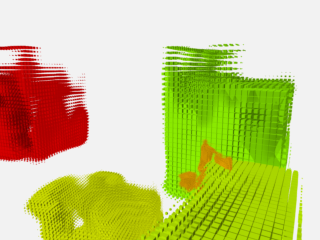}}
    & \frame{\includegraphics[width=0.12\textwidth]{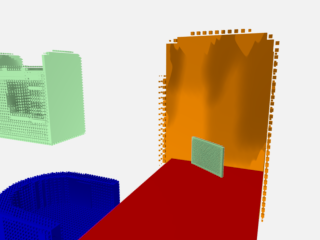}}
    & \frame{\includegraphics[width=0.12\textwidth]{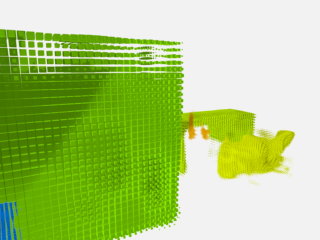}}
    & \frame{\includegraphics[width=0.12\textwidth]{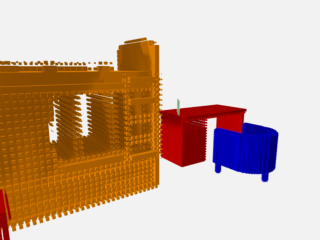}}
    & \frame{\includegraphics[width=0.12\textwidth]{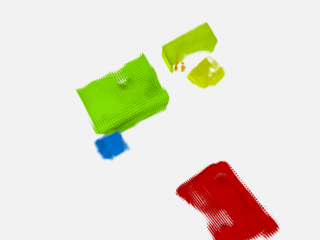}}
    & \frame{\includegraphics[width=0.12\textwidth]{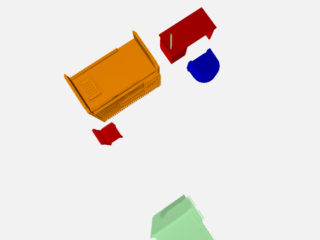}}\\

    \frame{\includegraphics[width=0.12\textwidth]{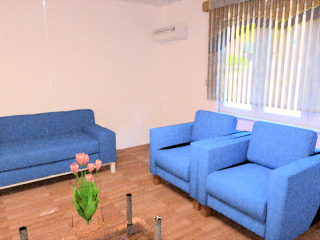}}
    & \frame{\includegraphics[width=0.12\textwidth]{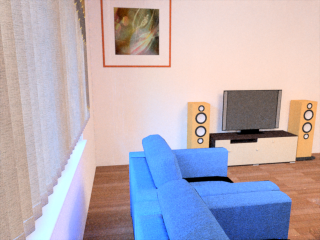}}
    & \frame{\includegraphics[width=0.12\textwidth]{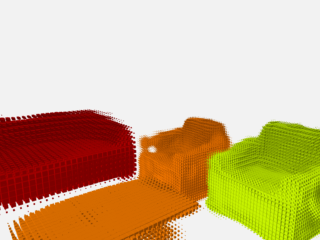}}
    & \frame{\includegraphics[width=0.12\textwidth]{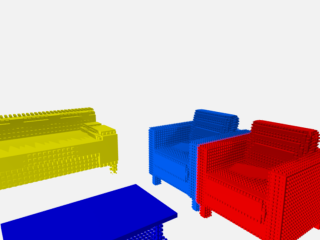}}
    & \frame{\includegraphics[width=0.12\textwidth]{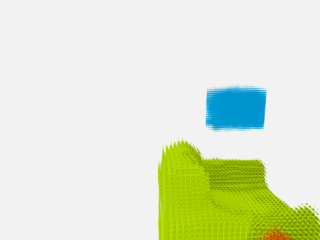}}
    & \frame{\includegraphics[width=0.12\textwidth]{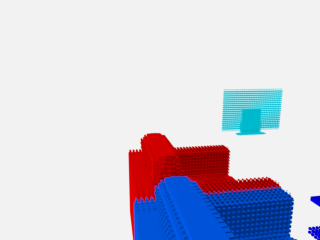}}
    & \frame{\includegraphics[width=0.12\textwidth]{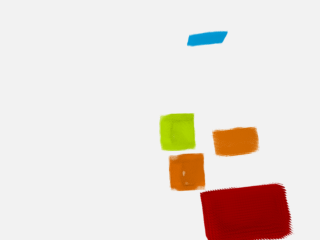}}
    & \frame{\includegraphics[width=0.12\textwidth]{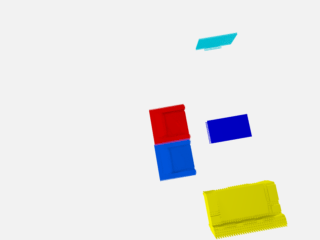}}\\
    
    \frame{\includegraphics[width=0.12\textwidth]{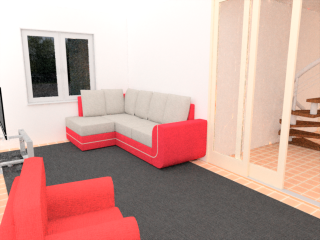}}
    & \frame{\includegraphics[width=0.12\textwidth]{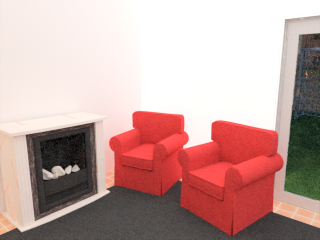}}
    & \frame{\includegraphics[width=0.12\textwidth]{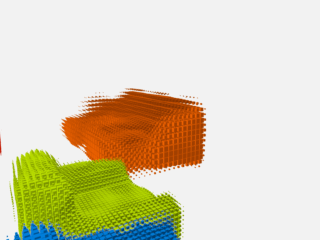}}
    & \frame{\includegraphics[width=0.12\textwidth]{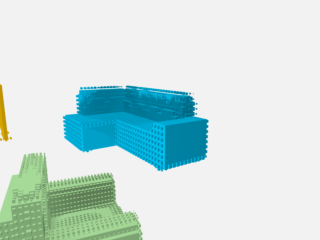}}
    & \frame{\includegraphics[width=0.12\textwidth]{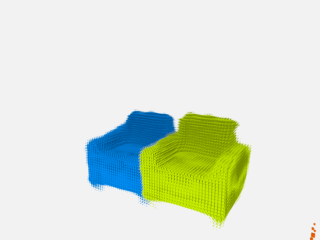}}
    & \frame{\includegraphics[width=0.12\textwidth]{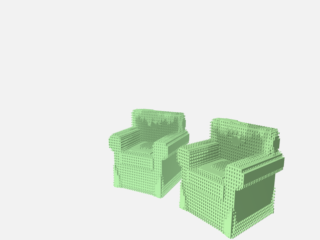}}
    & \frame{\includegraphics[width=0.12\textwidth]{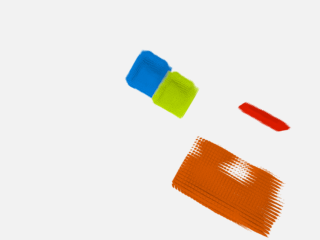}}
    & \frame{\includegraphics[width=0.12\textwidth]{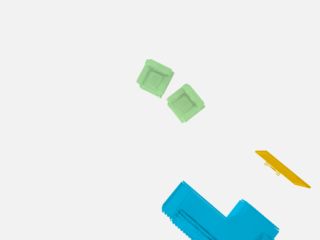}}\\

    \frame{\includegraphics[width=0.12\textwidth]{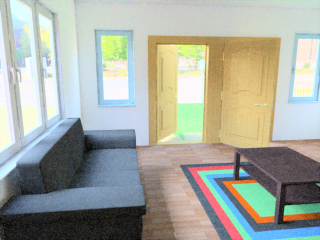}}
    & \frame{\includegraphics[width=0.12\textwidth]{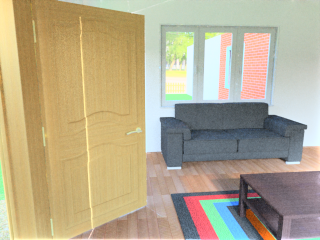}}
    & \frame{\includegraphics[width=0.12\textwidth]{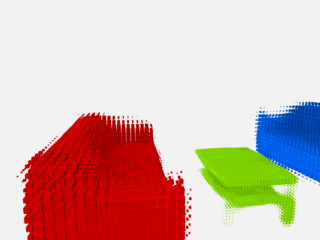}}
    & \frame{\includegraphics[width=0.12\textwidth]{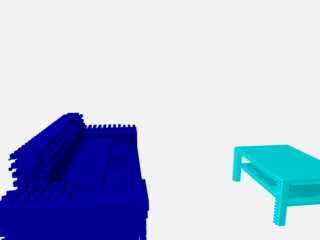}}
    & \frame{\includegraphics[width=0.12\textwidth]{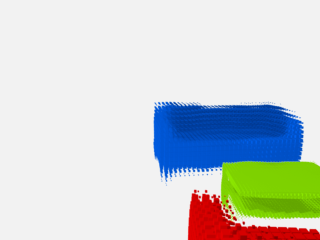}}
    & \frame{\includegraphics[width=0.12\textwidth]{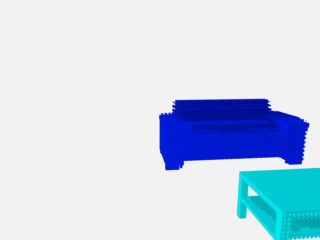}}
    & \frame{\includegraphics[width=0.12\textwidth]{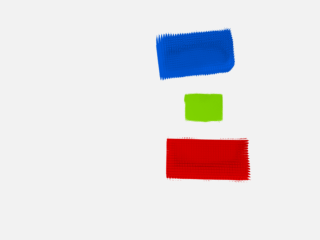}}
    & \frame{\includegraphics[width=0.12\textwidth]{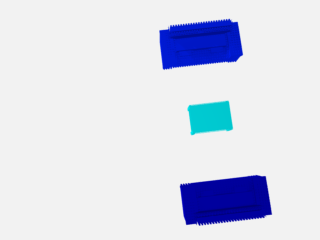}}\\
    
    \frame{\includegraphics[width=0.12\textwidth]{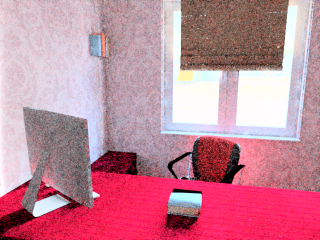}}
    & \frame{\includegraphics[width=0.12\textwidth]{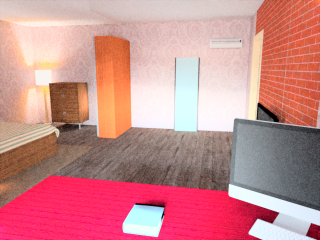}}
    & \frame{\includegraphics[width=0.12\textwidth]{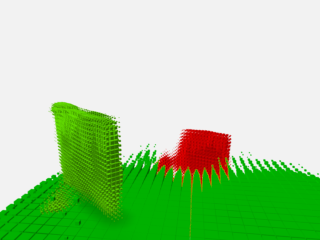}}
    & \frame{\includegraphics[width=0.12\textwidth]{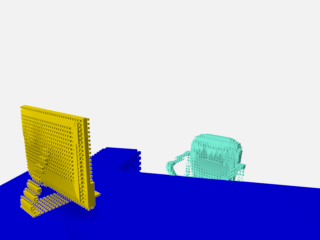}}
    & \frame{\includegraphics[width=0.12\textwidth]{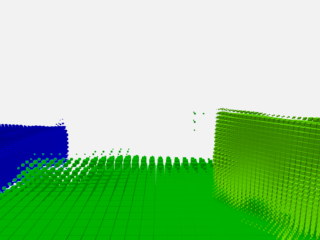}}
    & \frame{\includegraphics[width=0.12\textwidth]{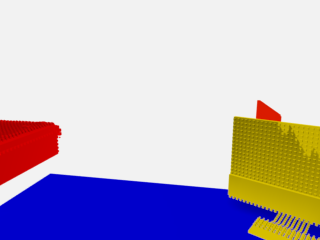}}
    & \frame{\includegraphics[width=0.12\textwidth]{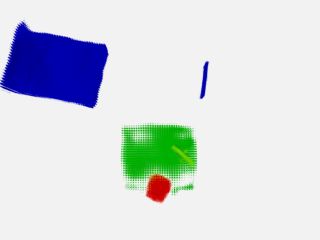}}
    & \frame{\includegraphics[width=0.12\textwidth]{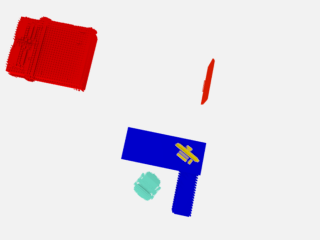}}\\
    
    \frame{\includegraphics[width=0.12\textwidth]{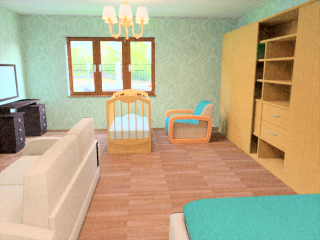}}
    & \frame{\includegraphics[width=0.12\textwidth]{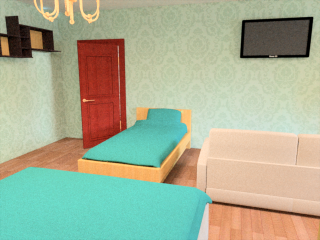}}
    & \frame{\includegraphics[width=0.12\textwidth]{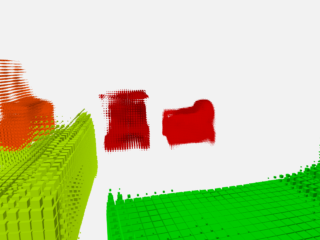}}
    & \frame{\includegraphics[width=0.12\textwidth]{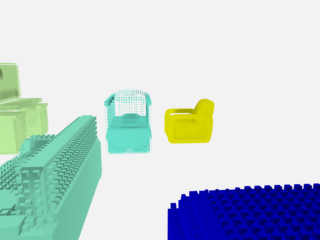}}
    & \frame{\includegraphics[width=0.12\textwidth]{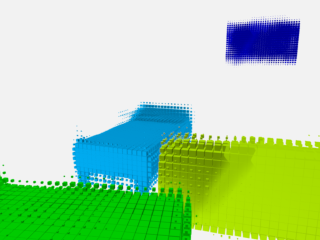}}
    & \frame{\includegraphics[width=0.12\textwidth]{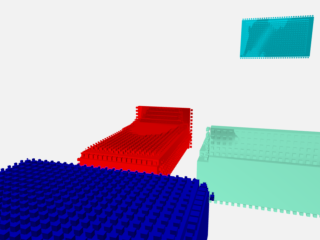}}
    & \frame{\includegraphics[width=0.12\textwidth]{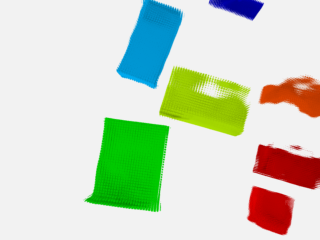}}
    & \frame{\includegraphics[width=0.12\textwidth]{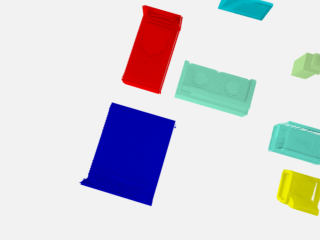}}\\

    \frame{\includegraphics[width=0.12\textwidth]{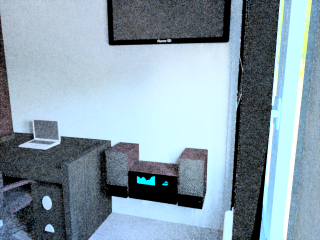}}
    & \frame{\includegraphics[width=0.12\textwidth]{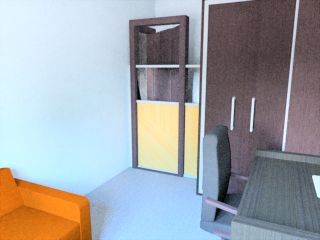}}
    & \frame{\includegraphics[width=0.12\textwidth]{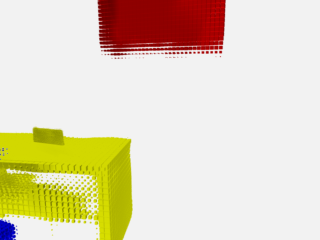}}
    & \frame{\includegraphics[width=0.12\textwidth]{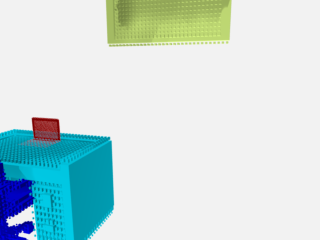}}
    & \frame{\includegraphics[width=0.12\textwidth]{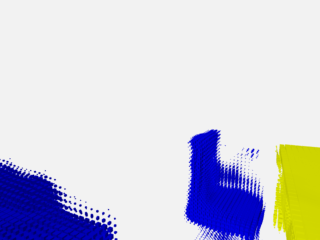}}
    & \frame{\includegraphics[width=0.12\textwidth]{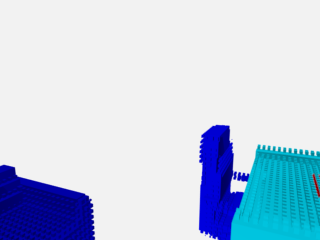}}
    & \frame{\includegraphics[width=0.12\textwidth]{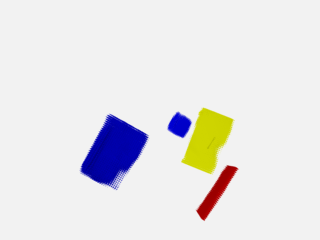}}
    & \frame{\includegraphics[width=0.12\textwidth]{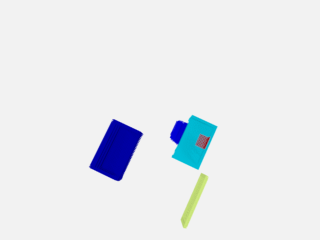}}\\

    \bottomrule
    \end{tabular}
    \caption{Additional Qualitative results on the SUNCG test set. The final 3D predictions are shown in three different camera poses (1) the same camera as image 1; (2) the same camera as image 2; (3) a bird view to see all the objects in the whole scene. In the prediction, red/orange objects are from the left image, blue objects are from the right image, green/yellow objects are stitched from both the images.}
    \label{fig:suncg-additional}
\end{figure*}

We show additional qualitative examples in 
Fig. \ref{fig:suncg-additional}, \ref{fig:comparison}, and \ref{fig:stitching}
which follows the same format as Fig. 3, 4 and 5 of the main paper.
For better visualization, we also put a video into our supplementary material.

\begin{figure}[t]
    \centering
    \scriptsize
    \begin{tabular}{c@{\hskip3pt}c@{\hskip8pt}c@{\hskip3pt}c@{\hskip3pt}c@{\hskip3pt}c@{\hskip8pt}c}
    \toprule
    
    Image 1 & Image 2 & Feedforward & NMS & Raw Affinity & {\bf \papername} & GT\\
    \midrule

    \frame{\includegraphics[width=0.13\textwidth]{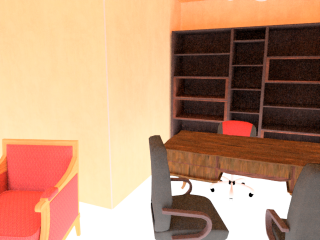}}
    & \frame{\includegraphics[width=0.13\textwidth]{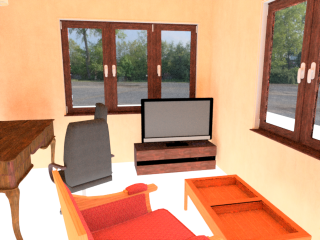}}
    & \frame{\includegraphics[width=0.13\textwidth]{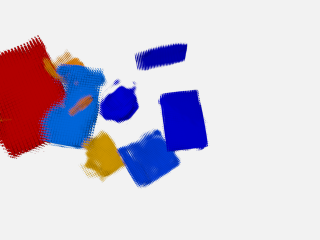}}
    & \frame{\includegraphics[width=0.13\textwidth]{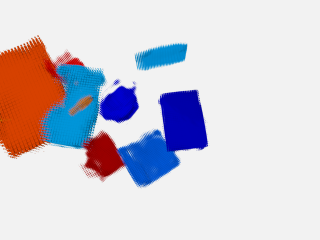}}
    & \frame{\includegraphics[width=0.13\textwidth]{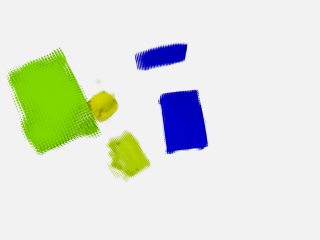}}
    & \frame{\includegraphics[width=0.13\textwidth]{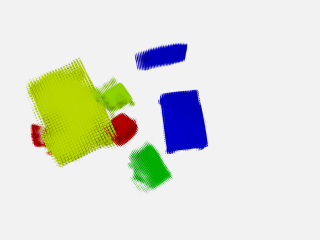}}
    & \frame{\includegraphics[width=0.13\textwidth]{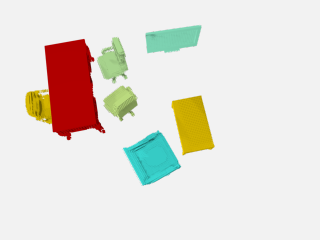}}\\

    \frame{\includegraphics[width=0.13\textwidth]{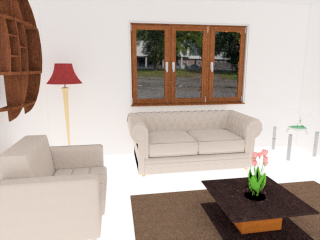}}
    & \frame{\includegraphics[width=0.13\textwidth]{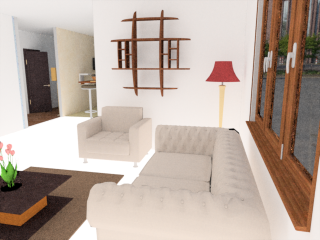}}
    & \frame{\includegraphics[width=0.13\textwidth]{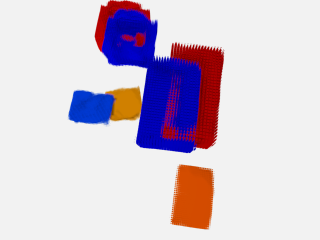}}
    & \frame{\includegraphics[width=0.13\textwidth]{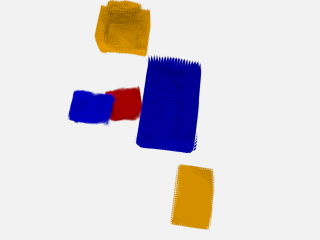}}
    & \frame{\includegraphics[width=0.13\textwidth]{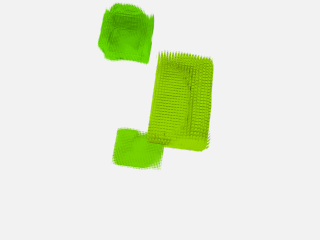}}
    & \frame{\includegraphics[width=0.13\textwidth]{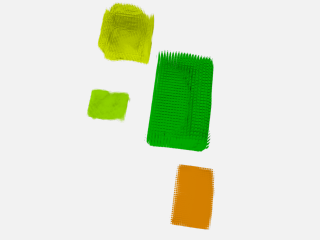}}
    & \frame{\includegraphics[width=0.13\textwidth]{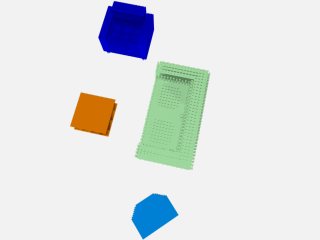}}\\

    \frame{\includegraphics[width=0.13\textwidth]{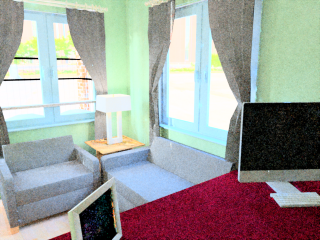}}
    & \frame{\includegraphics[width=0.13\textwidth]{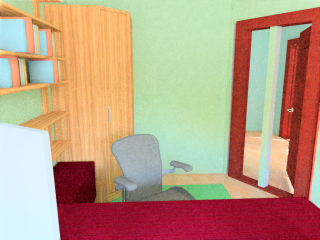}}
    & \frame{\includegraphics[width=0.13\textwidth]{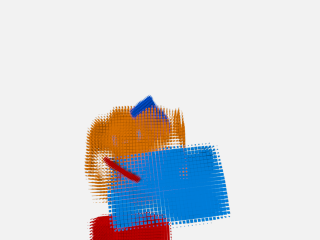}}
    & \frame{\includegraphics[width=0.13\textwidth]{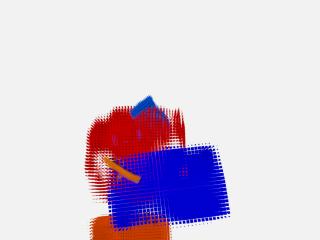}}
    & \frame{\includegraphics[width=0.13\textwidth]{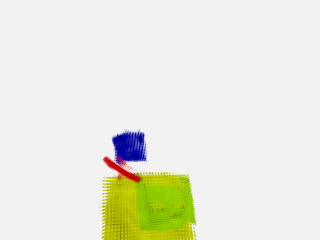}}
    & \frame{\includegraphics[width=0.13\textwidth]{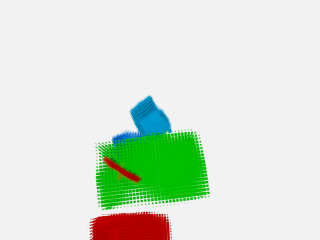}}
    & \frame{\includegraphics[width=0.13\textwidth]{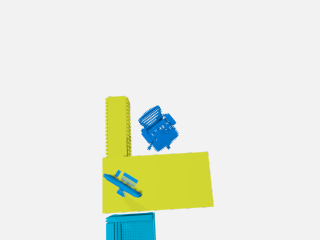}}\\

    \frame{\includegraphics[width=0.13\textwidth]{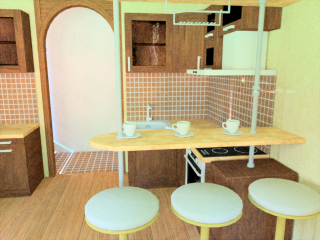}}
    & \frame{\includegraphics[width=0.13\textwidth]{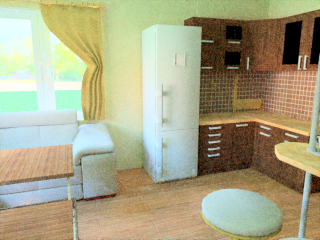}}
    & \frame{\includegraphics[width=0.13\textwidth]{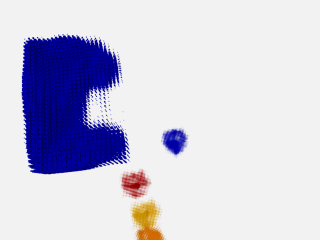}}
    & \frame{\includegraphics[width=0.13\textwidth]{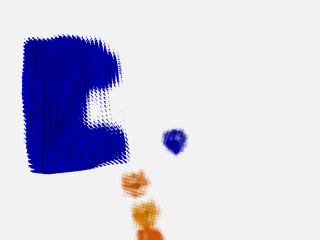}}
    & \frame{\includegraphics[width=0.13\textwidth]{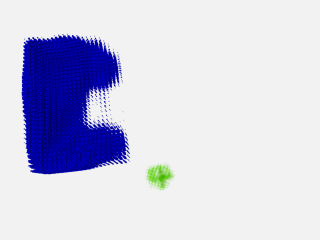}}
    & \frame{\includegraphics[width=0.13\textwidth]{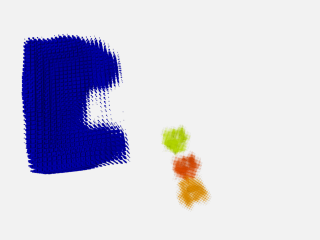}}
    & \frame{\includegraphics[width=0.13\textwidth]{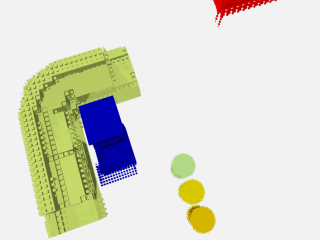}}\\

    \frame{\includegraphics[width=0.13\textwidth]{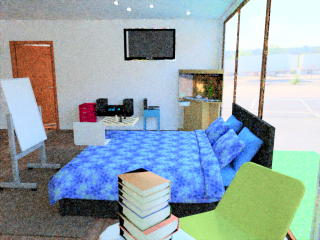}}
    & \frame{\includegraphics[width=0.13\textwidth]{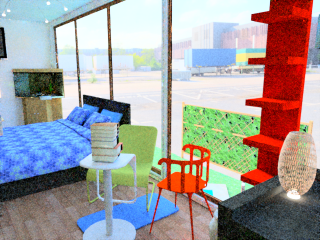}}
    & \frame{\includegraphics[width=0.13\textwidth]{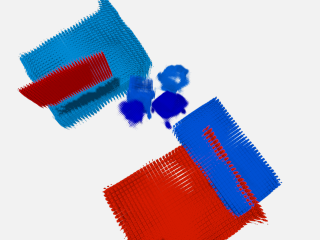}}
    & \frame{\includegraphics[width=0.13\textwidth]{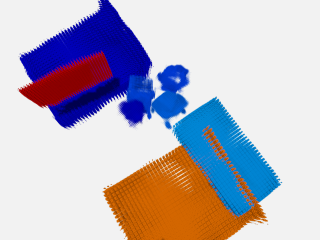}}
    & \frame{\includegraphics[width=0.13\textwidth]{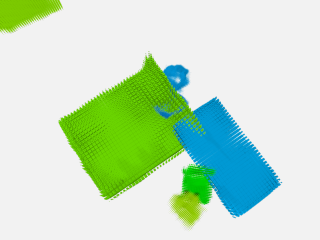}}
    & \frame{\includegraphics[width=0.13\textwidth]{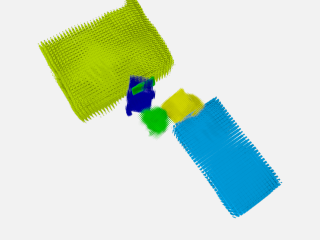}}
    & \frame{\includegraphics[width=0.13\textwidth]{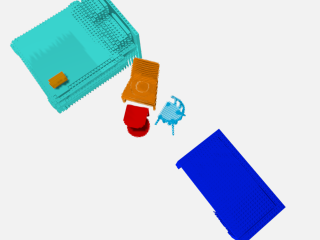}}\\

    \frame{\includegraphics[width=0.13\textwidth]{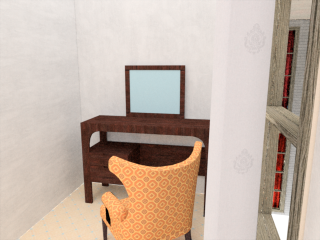}}
    & \frame{\includegraphics[width=0.13\textwidth]{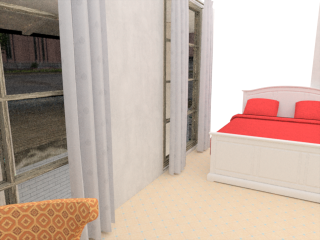}}
    & \frame{\includegraphics[width=0.13\textwidth]{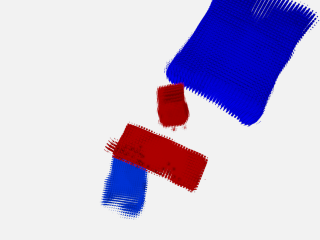}}
    & \frame{\includegraphics[width=0.13\textwidth]{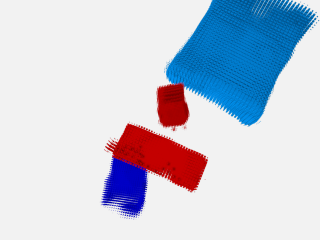}}
    & \frame{\includegraphics[width=0.13\textwidth]{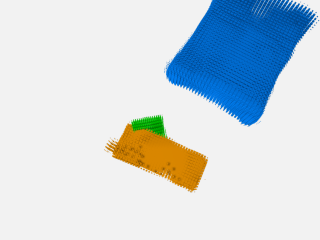}}
    & \frame{\includegraphics[width=0.13\textwidth]{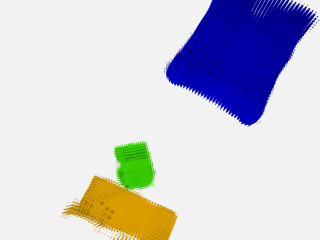}}
    & \frame{\includegraphics[width=0.13\textwidth]{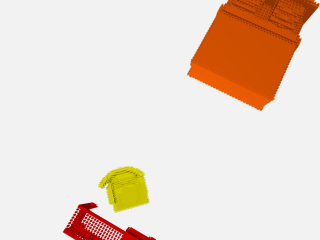}}\\

    \frame{\includegraphics[width=0.13\textwidth]{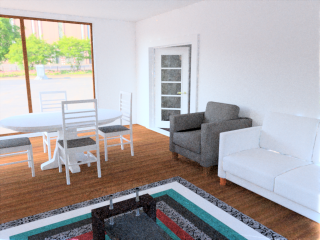}}
    & \frame{\includegraphics[width=0.13\textwidth]{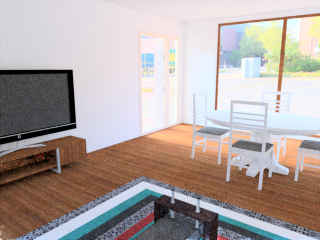}}
    & \frame{\includegraphics[width=0.13\textwidth]{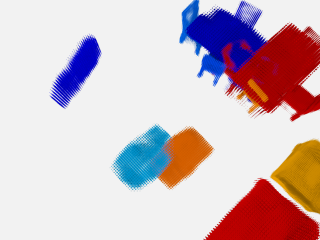}}
    & \frame{\includegraphics[width=0.13\textwidth]{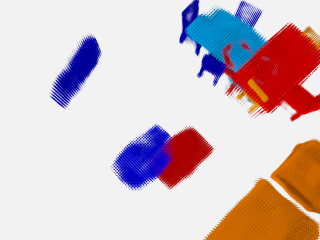}}
    & \frame{\includegraphics[width=0.13\textwidth]{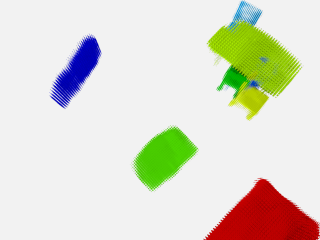}}
    & \frame{\includegraphics[width=0.13\textwidth]{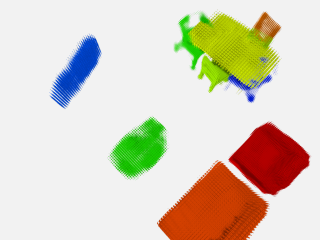}}
    & \frame{\includegraphics[width=0.13\textwidth]{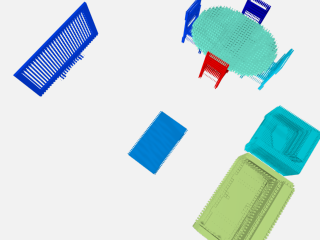}}\\

    \frame{\includegraphics[width=0.13\textwidth]{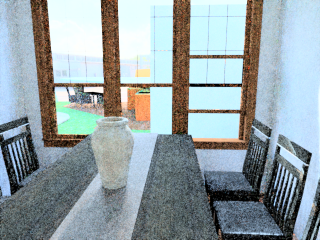}}
    & \frame{\includegraphics[width=0.13\textwidth]{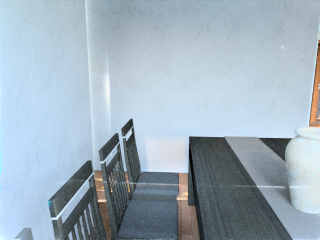}}
    & \frame{\includegraphics[width=0.13\textwidth]{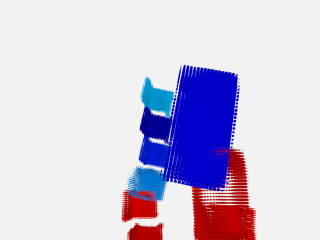}}
    & \frame{\includegraphics[width=0.13\textwidth]{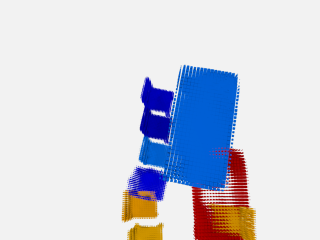}}
    & \frame{\includegraphics[width=0.13\textwidth]{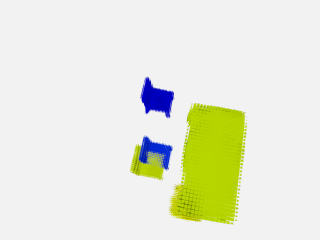}}
    & \frame{\includegraphics[width=0.13\textwidth]{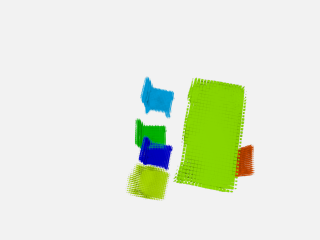}}
    & \frame{\includegraphics[width=0.13\textwidth]{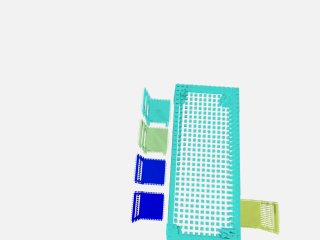}}\\

    \frame{\includegraphics[width=0.13\textwidth]{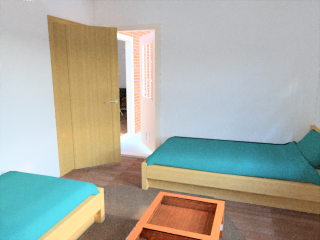}}
    & \frame{\includegraphics[width=0.13\textwidth]{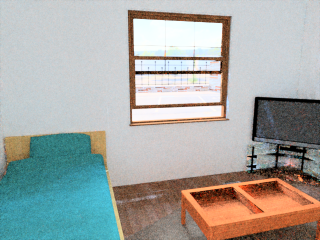}}
    & \frame{\includegraphics[width=0.13\textwidth]{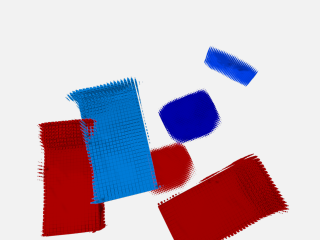}}
    & \frame{\includegraphics[width=0.13\textwidth]{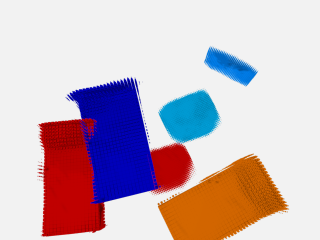}}
    & \frame{\includegraphics[width=0.13\textwidth]{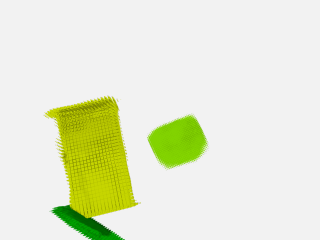}}
    & \frame{\includegraphics[width=0.13\textwidth]{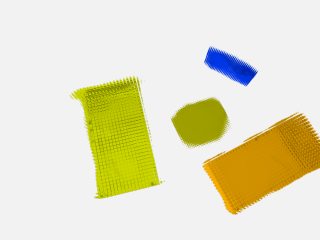}}
    & \frame{\includegraphics[width=0.13\textwidth]{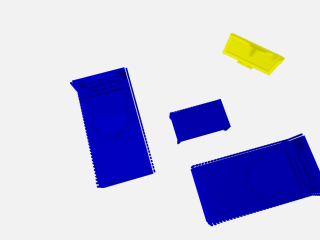}}\\

    \frame{\includegraphics[width=0.13\textwidth]{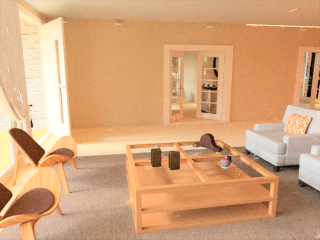}}
    & \frame{\includegraphics[width=0.13\textwidth]{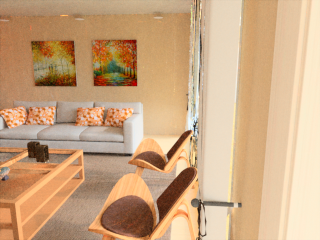}}
    & \frame{\includegraphics[width=0.13\textwidth]{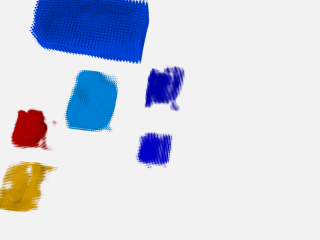}}
    & \frame{\includegraphics[width=0.13\textwidth]{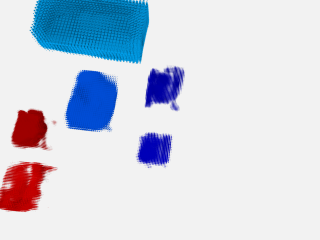}}
    & \frame{\includegraphics[width=0.13\textwidth]{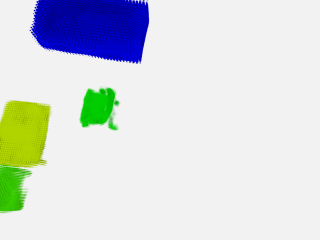}}
    & \frame{\includegraphics[width=0.13\textwidth]{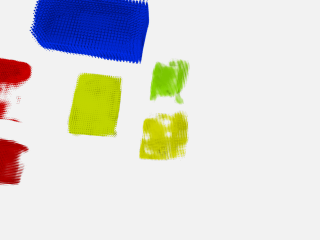}}
    & \frame{\includegraphics[width=0.13\textwidth]{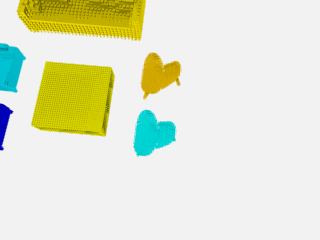}}\\

    \bottomrule
    \end{tabular}
    \caption{Comparison between {\papername} and alternative approaches. All outputs are shown from bird's eye view.}
    \label{fig:comparison}
\end{figure}

\begin{figure}[t]
    \centering
    
    \begin{tabular}{c@{\hskip3pt}c@{\hskip8pt}c@{\hskip3pt}c@{\hskip8pt}c@{\hskip3pt}c}
    \toprule
    
    Before & After & Before & After & Before & After\\
    \midrule
    \includegraphics[width=0.14\textwidth]{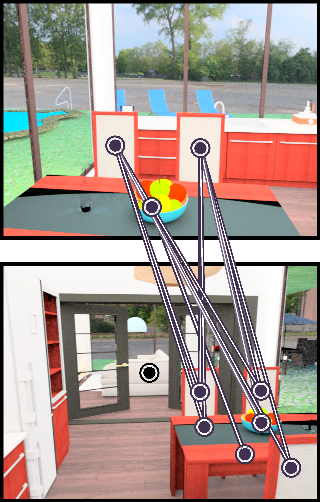}
    & \includegraphics[width=0.14\textwidth]{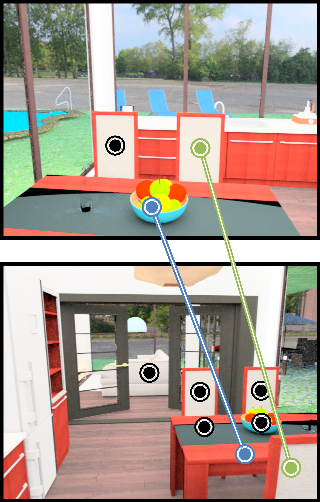}
    & \includegraphics[width=0.14\textwidth]{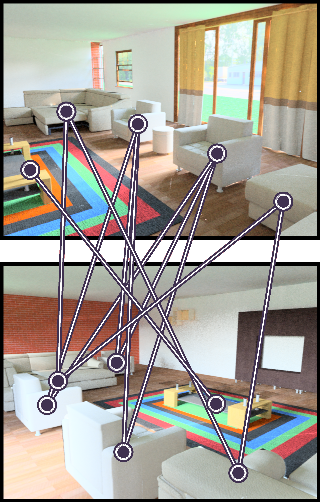}
    & \includegraphics[width=0.14\textwidth]{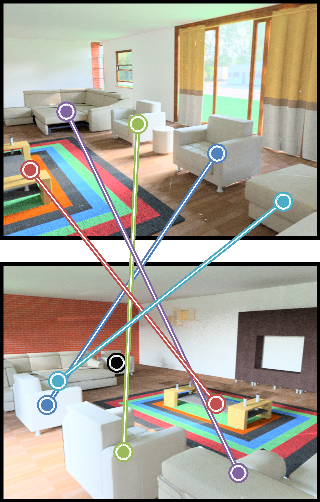}
    & \includegraphics[width=0.14\textwidth]{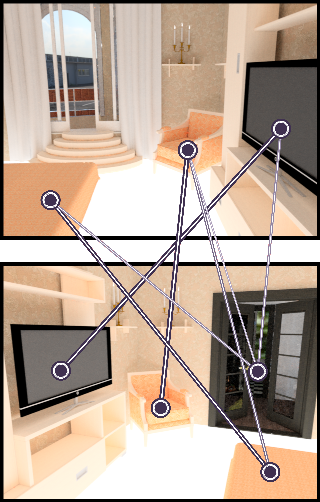}
    & \includegraphics[width=0.14\textwidth]{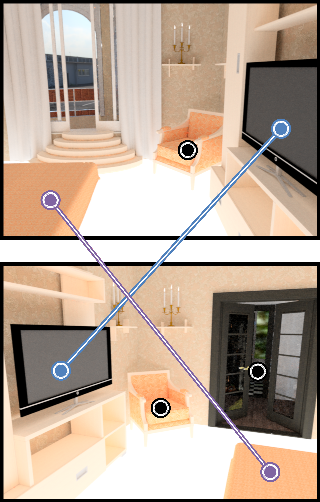}\\
    
    \includegraphics[width=0.14\textwidth]{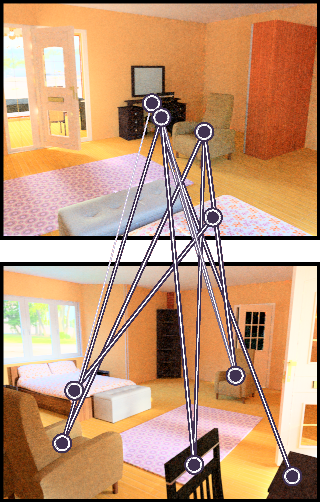}
    & \includegraphics[width=0.14\textwidth]{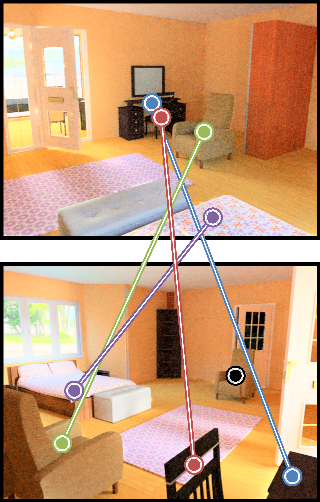}
    & \includegraphics[width=0.14\textwidth]{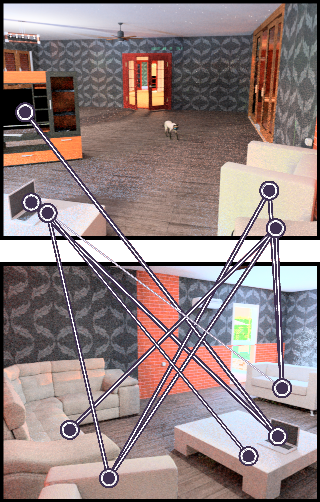}
    & \includegraphics[width=0.14\textwidth]{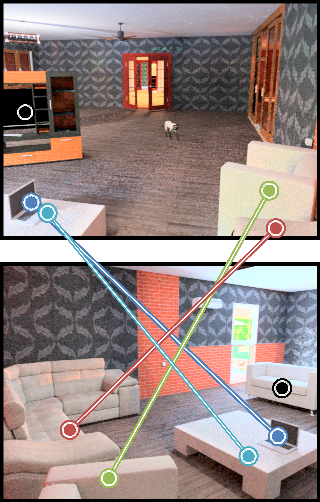}
    & \includegraphics[width=0.14\textwidth]{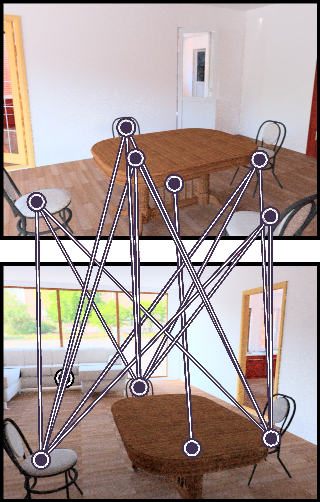}
    & \includegraphics[width=0.14\textwidth]{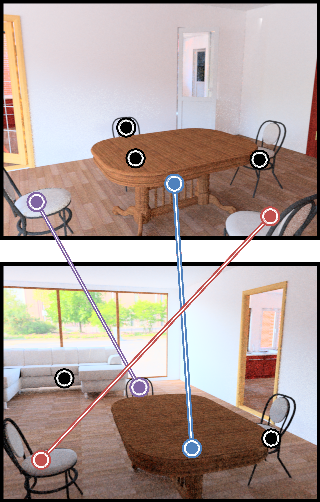}\\

    \includegraphics[width=0.14\textwidth]{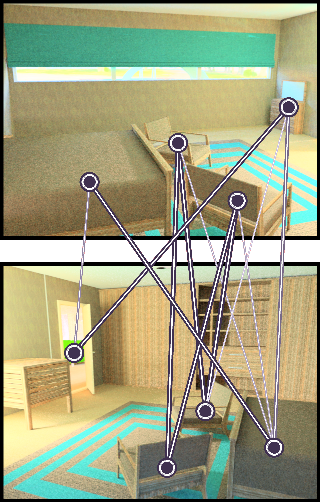}
    & \includegraphics[width=0.14\textwidth]{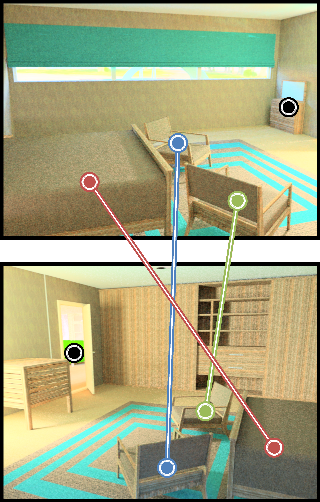}
    & \includegraphics[width=0.14\textwidth]{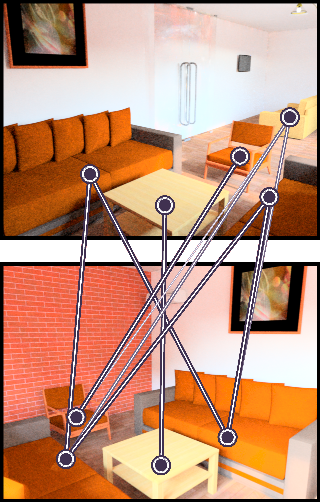}
    & \includegraphics[width=0.14\textwidth]{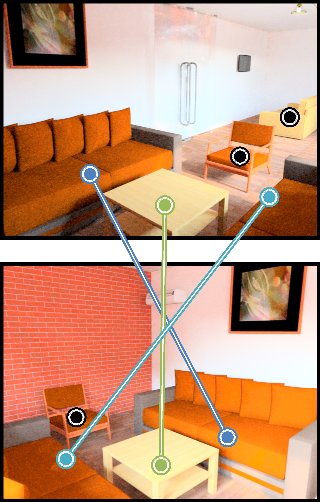}
    & \includegraphics[width=0.14\textwidth]{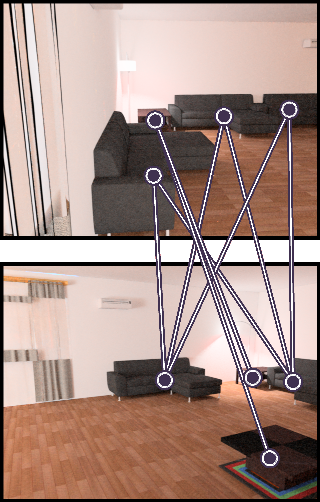}
    & \includegraphics[width=0.14\textwidth]{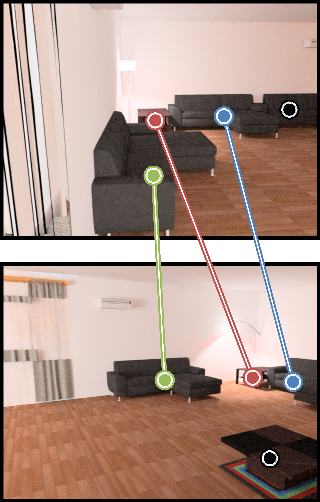}\\

    \includegraphics[width=0.14\textwidth]{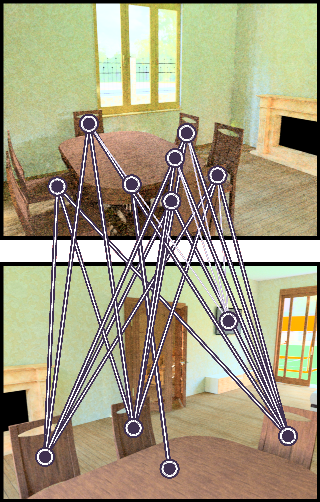}
    & \includegraphics[width=0.14\textwidth]{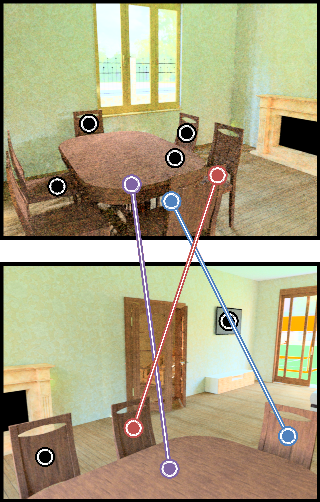}
    & \includegraphics[width=0.14\textwidth]{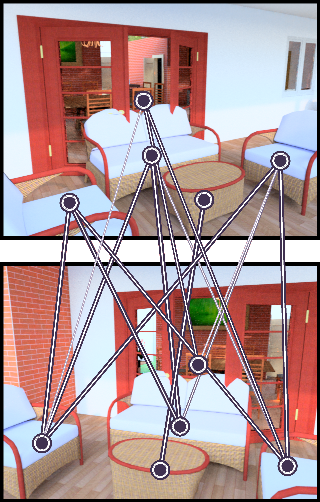}
    & \includegraphics[width=0.14\textwidth]{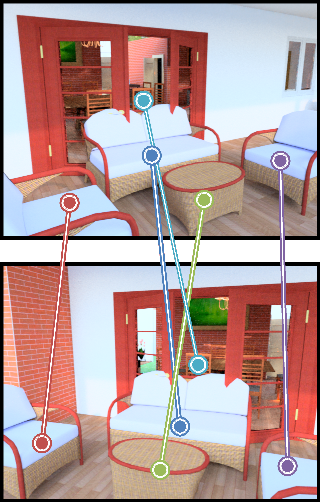}
    & \includegraphics[width=0.14\textwidth]{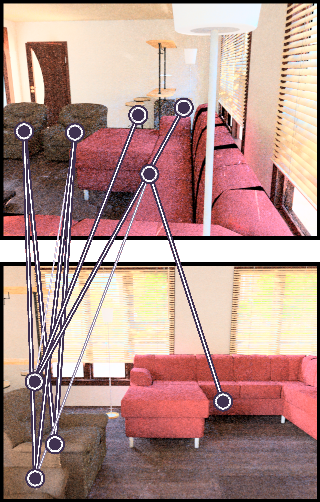}
    & \includegraphics[width=0.14\textwidth]{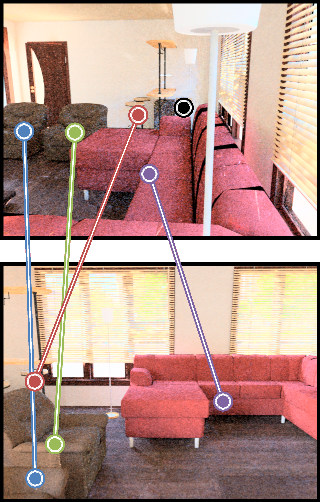}\\
    \includegraphics[width=0.14\textwidth]{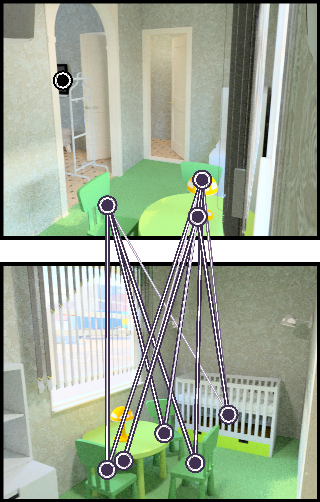}
    & \includegraphics[width=0.14\textwidth]{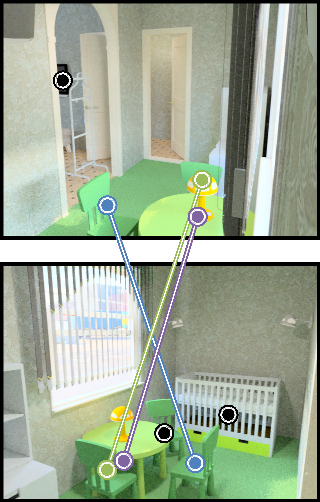}
    & \includegraphics[width=0.14\textwidth]{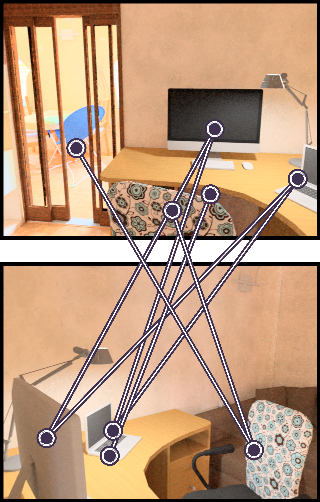}
    & \includegraphics[width=0.14\textwidth]{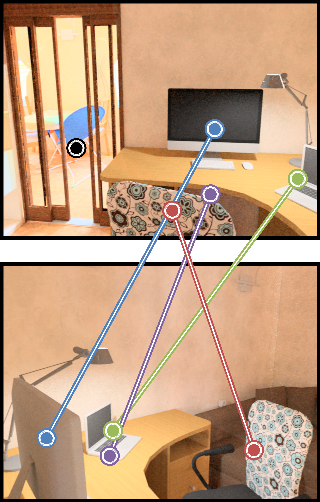}
    & \includegraphics[width=0.14\textwidth]{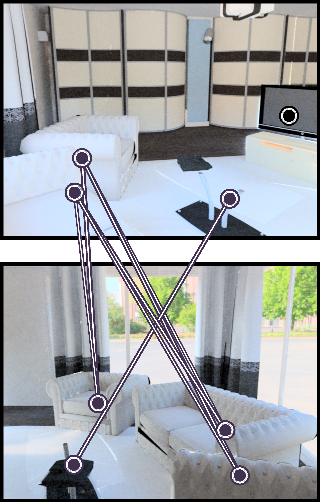}
    & \includegraphics[width=0.14\textwidth]{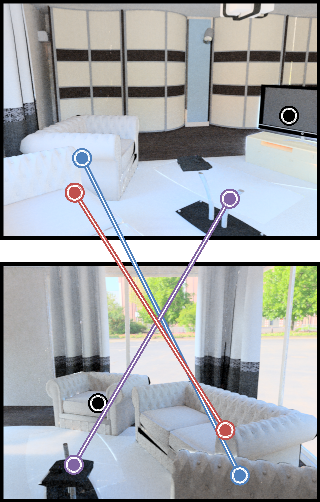}\\

    \bottomrule
    \end{tabular}
    \caption{Visualization of the stitching stage. The affinity matrix generates proposals of corresponding objects, and then the stitching stage removes outliers by inferring the most likely explanation of the scene. Before stitching, the thickness and darkness of the line represent the value of affinity score. The thicker / darker, the higher the value in the affinity matrix. }
    \label{fig:stitching}
\end{figure} 
\end{document}